%% file: main.tex
\documentclass[10pt,twocolumn]{article}
\usepackage[left=0.65in,right=0.65in,top=0.75in,bottom=0.75in]{geometry}
\usepackage[hyphens]{url}
\usepackage{graphicx}
\usepackage[T1]{fontenc}
\usepackage{lmodern}
\usepackage{microtype}
\usepackage[authoryear,round]{natbib}
\usepackage{caption}
\usepackage{booktabs}
\usepackage{adjustbox}
\usepackage{tabularx}
\usepackage{stfloats}
\usepackage{amsmath,amssymb,amsfonts}
\usepackage{tikz}
\usetikzlibrary{arrows.meta,backgrounds,calc,fit,positioning}
\usepackage[hidelinks]{hyperref}
\graphicspath{{figures/}}
\title{Multi-View 3D Reconstruction Using V-JEPA Features\thanks{Source code: \url{https://github.com/Anwarshamim01/multi-view-3d-reconstruction-using-v-jepa-features}.}}
\hypersetup{pdftitle={Multi-View 3D Reconstruction Using V-JEPA Features}}

\author{Anwar Shamim \and Jeffrin Sam}
\date{}

\begin{document}

\maketitle

\begin{abstract}
\input{chapters/abstract}
\end{abstract}

\input{chapters/introduction}
\input{chapters/related_work}
\input{chapters/problem_formulation}
\input{chapters/architecture}
\input{chapters/learning_objectives}
\input{chapters/experimental_protocol}
\input{chapters/results}
\input{chapters/discussion}

\clearpage
\small
\bibliographystyle{abbrvnat}
\bibliography{references}
\normalsize

\clearpage
\appendix
\input{appendices/architecture_implementation}
\input{appendices/objective_details}
\input{appendices/data_protocol_reproducibility}
\input{appendices/metric_definitions}
\input{appendices/detailed_results}
\input{appendices/surface_render_tables}
\clearpage
\input{appendices/visual_results}
\clearpage
\input{chapters/additional_experiments}

\end{document}

%% file: chapters/abstract.tex
Predictive video encoders provide dense visual representations, but
adapting them to reconstruction requires jointly accurate depth,
cameras and cross-view geometry. We investigate this problem using
a supervised RGB-only multi-view architecture that combines
video-pretrained features, camera estimation, scene aggregation,
epipolar transport and residual pointmaps. Experiments compare global
camera regression, spatial correspondence reasoning and joint relative
objectives on CO3D, alongside an independently trained TUM RGB-D model.
On 168 CO3D five-view windows, the spatial-camera configuration attains
29.3\% lower absolute trajectory error than the global-camera configuration,
but higher trajectory error than identity poses and higher depth
absolute relative error than the global-camera configuration.
Compared with the spatial-camera configuration, the joint
relative-objective configuration has lower normalized
intrinsics error and similarity-aligned Chamfer distance, from 0.6447
to 0.5573, but higher raw Chamfer distance, from 3.5535 to 4.4374
in dataset coordinates. Architecture, supervision and model selection
vary together, so these comparisons do not isolate component effects.
On 163 indoor windows, camera error is 0.3063\,m versus 0.3694\,m
for identity poses, whereas depth absolute relative error is 0.4279
versus 0.2516 for constant training-median depth. Refinement worsens
indoor camera error, and supplementary component experiments also exhibit
competing depth and pose effects. Confidence calibration and
latent-target retrieval further indicate that high predicted confidence
or latent cosine similarity alone does not establish geometric accuracy.
The findings distinguish
representation adaptation, aligned shape and raw reconstruction,
and identify the need for isolated component and encoder comparisons
before attributing a general reconstruction advantage to predictive
video pretraining.

%% file: chapters/introduction.tex
\section{Introduction}
\label{sec:introduction}

Reconstructing a scene from a few uncalibrated images requires more than
recognizing the objects they contain. A model must associate locations
across views, infer camera motion and calibration, and place surfaces in
a common coordinate system. Classical structure-from-motion estimates
these quantities through correspondences and geometric optimization
\citep{schoenberger2016sfm}. Learned reconstruction uses image-based
pointmap regression and matching, with systems such as VGGT also
directly predicting cameras, depth and tracks
\citep{wang2024dust3r,leroy2024mast3r,wang2025vggt}. This progress raises a
representation question: how much of the information learned through
video prediction can be adapted to explicit multi-view geometry?

Joint-Embedding Predictive Architectures learn by predicting latent
features rather than reconstructing pixels
\citep{assran2023ijepa,bardes2024vjepa,assran2025vjepa2}.
V-JEPA 2.1 extends this approach with dense and deep supervision
\citep{murlabadia2026vjepa21}. Such features provide a plausible
initialization for reconstruction, but dense representation quality
alone does not establish accurate metric depth, camera motion or
cross-view consistency. In particular, a low error after fitting a
reference-derived similarity transform can coexist with inaccurate raw
geometry in the first-view coordinate frame.

We investigate these distinctions with our reconstruction model,
initialized from a V-JEPA 2.1 encoder. Each input image is
encoded independently; a downstream camera model, scene-token
transformer and epipolar transport module provide multi-view reasoning.
A local RGB branch supplements the pretrained features, and a decoder
predicts depth, pointmaps, normals, confidence and support. A
camera-conditioned latent-view objective distills observed-view teacher
features, following the projection-conditioned prediction principle of
CrossJEPA \citep{perera2025crossjepa}. Because the target images are
observed, this objective concerns feature distillation rather than
prediction of an unseen future or withheld view.

The experiments address three questions. Does retaining spatial matches
improve camera estimation over global feature pooling? Do relative
geometry and calibration objectives improve depth and raw reconstruction
together? Does latent agreement or predicted confidence provide evidence
of useful geometric information? We train separately on object-centric
CO3D and indoor TUM RGB-D, and compare learned outputs with identity
cameras, constant training-median depth and training-median calibration.
These controls distinguish an improvement over another learned
configuration from an improvement over a simple geometric prior.

The answers are mixed. The spatial-camera configuration has lower CO3D
camera error than the global-camera configuration, but higher error
than identity poses on average. The relative-objective configuration
has lower calibration and aligned-shape errors but higher depth and
raw-surface errors. Architecture, supervision and selection criteria
change together, limiting causal attribution. On TUM, camera error is
lower than identity, but depth AbsRel exceeds constant depth in every
test window. Earlier component experiments also show
that a lower depth error can accompany a higher pose error. We therefore
frame the contribution as a method and an empirical characterization
of representation adaptation, rather than a claim of uniformly better
reconstruction.

Our contributions are: (i) a camera-conditioned architecture for adapting
dense video-pretrained features to supervised RGB-only reconstruction;
(ii) an evaluation that separates depth, camera, calibration and raw
versus similarity-aligned surfaces across two independently trained
settings; and (iii) an analysis of spatial camera reasoning, relative
objectives, confidence reliability and observed-view distillation,
including the limits of the available component experiments.

%% file: chapters/related_work.tex
\section{Related Work}

I-JEPA predicts masked image-region embeddings from visible context
\citep{assran2023ijepa}; V-JEPA extends latent prediction to video
\citep{bardes2024vjepa}, V-JEPA 2 scales video pretraining and adds
action-conditioned post-training \citep{assran2025vjepa2}, and V-JEPA
2.1 trains dense features at multiple encoder depths
\citep{murlabadia2026vjepa21}. We use a pretrained V-JEPA 2.1 encoder as
the visual backbone of the reconstruction system.

CrossJEPA transfers image-encoder representations to point clouds by
training a point-cloud encoder to predict features of rendered views,
conditioned on view parameters; target embeddings are cached to reduce
repeated teacher computation \citep{perera2025crossjepa}. Our
geometry-conditioned latent-view prediction (G-LVP) objective has a
related structure: it uses
camera-conditioned features from observed views to predict cached
V-JEPA embeddings of target views. The setting differs in both input
representation and task: G-LVP is an auxiliary objective in a
supervised image-based reconstruction model, rather than a point-cloud
pretraining objective.

Feed-forward reconstruction methods predict dense geometry and camera
information from images. DUSt3R regresses pixel-aligned pointmaps in a
shared frame \citep{wang2024dust3r}; MASt3R adds dense descriptors and
matching \citep{leroy2024mast3r}; and VGGT jointly predicts cameras,
depth, pointmaps, and tracks across multiple views
\citep{wang2025vggt}. Fast3R jointly processes many images in one forward
pass \citep{yang2025fast3r}; MUSt3R extends dense reconstruction to
multi-view inputs \citep{cabon2025must3r}; and SLAM3R incrementally
reconstructs dense scenes from RGB video \citep{liu2025slam3r}.
More recent formulations also question the choice of coordinate
reference and prediction targets. $\pi^3$ uses permutation-equivariant
processing without a designated reference view
\citep{wang2026pi3}; Depth Anything 3 jointly predicts depth and camera-ray
maps from arbitrary views \citep{lin2026depthanything3}. Our architecture
retains a first-view reference and explicit cameras with residual
pointmaps. Our study focuses on adapting a video-pretrained representation
under a limited training budget. These design differences motivate the
study; they do not imply superior accuracy or efficiency. A quantitative
comparison requires a common split, input resolution, view count, and
alignment protocol, which the present experiments do not provide.

Neural radiance fields optimize scene representations from calibrated
views \citep{mildenhall2020nerf}, while 3D Gaussian Splatting uses
explicit primitives for differentiable rendering
\citep{kerbl2023gaussians}. FreeSplatter predicts pixel-aligned Gaussians
from sparse uncalibrated images and recovers camera parameters using
geometric solvers \citep{xu2025freesplatter}.
Here, Gaussian rendering supplies a training
signal through a depth-derived scene representation, with attributes
constructed from image and geometry predictions. The evaluation
focuses on geometry rather than novel-view image quality. CO3D contains
object-centric videos with estimated cameras, masks, and depth
\citep{reizenstein2021co3d}. TUM RGB-D provides synchronized indoor RGB,
registered depth, and motion-capture trajectories
\citep{sturm2012tum}; the two datasets therefore test distinct
object-centric and metrically scaled indoor settings.

%% file: chapters/problem_formulation.tex
\section{Reconstruction Model}
\label{sec:model}
\subsection{Inputs, Outputs and Coordinate Reference}
\label{sec:formulation}

Given $K$ RGB images $\{I_k\}_{k=1}^{K}$, the model predicts intrinsics
$\hat A_k$, camera-to-world transforms $\hat T_k\in SE(3)$, depth
$\hat D_k$, unit normals $\hat N_k$, pointmaps $\hat X_k$, confidence
$\hat C_k$, support probability $\hat M_k$ and dense descriptors.
We use $A$ for the intrinsic matrix to distinguish it from the view
count $K$. The first camera defines the coordinate frame:
$\hat T_1=I_4$. Reference poses are transformed into the same frame
before supervision. Inference requires RGB images alone; depth,
intrinsics, masks and poses are training labels.

We distinguish unaligned and aligned reconstruction. Raw geometry
denotes the predicted pointmaps fused in the first-view coordinate
frame, retaining errors in shape, camera placement and learned scale.
Aligned geometry measures shape after fitting a least-squares similarity
transform \citep{umeyama1991least} to corresponding reference points,
\begin{equation}
\label{eq:similarity}
 (s^*,R^*,t^*)=\arg\min_{s>0,R\in SO(3),t}
 \sum_p\|sR\hat x_p+t-x_p\|_2^2.
\end{equation}
Here $\hat x_p,x_p\in\mathbb R^3$ are corresponding predicted and reference
points in that common frame, $s$ is scale and $t\in\mathbb R^3$ is
translation. In surface evaluation, this reference-derived transform
is a fitting operation rather than an RGB-only model output.
Its fitted scale, rotation and translation can
correct errors that remain relevant to camera-consistent reconstruction.

%% file: chapters/architecture.tex
\begin{figure*}[t]
\centering
\resizebox{0.98\textwidth}{!}{%
\begin{tikzpicture}[font=\sffamily\small,>=Latex,
box/.style={draw,rounded corners,align=center,minimum height=1.0cm,text width=2.7cm,fill=blue!5},
arr/.style={->,thick}]
\node[box] (rgb) at (0,0) {RGB images};
\node[box] (enc) at (3.4,0) {Independent view\\encoding};
\node[box] (cam) at (6.8,1.5) {Global or spatial\\camera estimation};
\node[box] (agg) at (10.2,0) {Scene aggregation\\and pose refinement};
\node[box] (epi) at (13.6,0) {Epipolar transport\\and dense decoder};
\node[box] (geo) at (17,0) {Depth, pointmaps,\\support, descriptors};
\node[box,fill=orange!10] (aux) at (6.8,-1.8) {Observed-view\\feature distillation};
\node[box] (edge) at (10.2,-3.2) {Local RGB pyramid};
\draw[arr] (rgb)--(enc);
\draw[arr] (enc)--(cam);
\draw[arr] (enc)--(agg);
\draw[arr] (cam)--(agg);
\draw[arr] (agg)--(epi);
\draw[arr] (epi)--(geo);
\draw[arr,dashed] (enc)--(aux);
\draw[arr,dashed] (cam)--(aux);
\draw[arr] (rgb.south) |- (edge.west);
\draw[arr] (edge.east) -| (epi.south);
\end{tikzpicture}}
\caption{Our reconstruction architecture. Video-pretrained features are extracted separately
from each image. Explicit camera estimation and epipolar transport
provide multi-view geometry. Dashed arrows denote a training-only
observed-view distillation objective. Cross-view rendering supervision
is omitted for clarity.}
\label{fig:architecture}
\end{figure*}
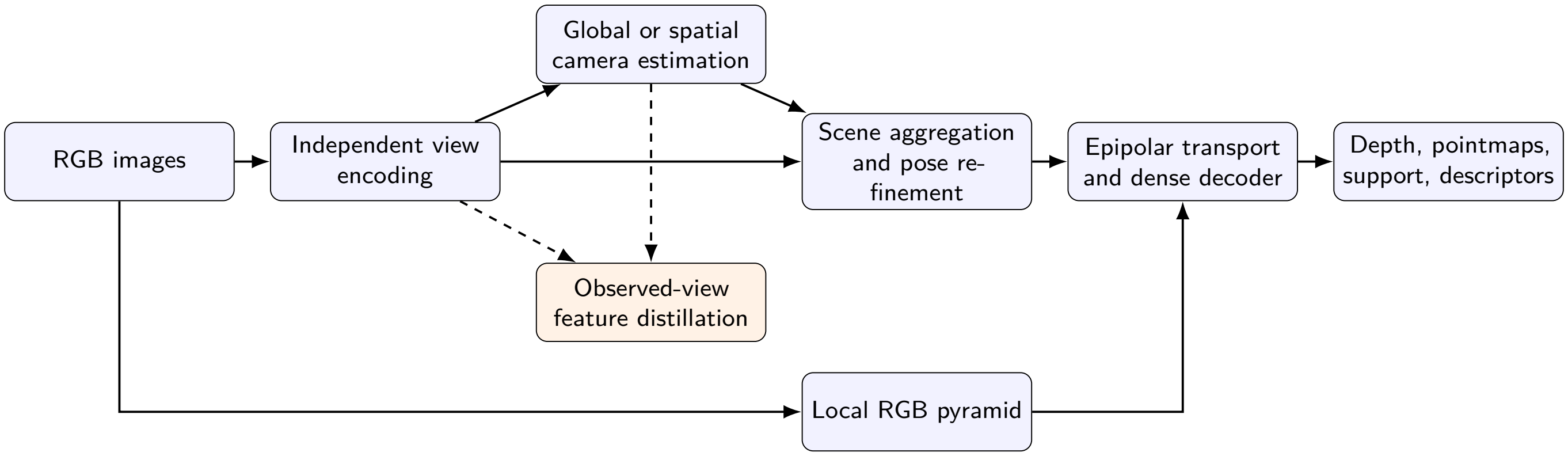

Figure~\ref{fig:architecture} summarizes the image features, camera
estimation and geometry prediction paths.

\subsection{Visual Features and Global Context}
A distilled V-JEPA 2.1 ViT-B/16 initializes the visual encoder
\citep{murlabadia2026vjepa21}. Two copies of each static image form a
two-frame clip, yielding a $32\times32$ grid of
768-dimensional tokens at $512\times512$ resolution. The repetitions
contain no motion. Thus video pretraining supplies the representation;
it does not jointly encode the reconstruction views as a video.
A convolutional RGB branch, evaluated at $384\times384$ pixels, supplies local
features to the dense decoder.

The model pools each view into a token and appends four learned scene
tokens. A two-layer, eight-head transformer exchanges context across
views. Initial camera estimates from Section~\ref{sec:camera} provide
embeddings of rotation, translation and normalized intrinsics.
The resulting view tokens modulate their spatial feature
maps through a learned channel gate, preserving the local feature grid.
Variable view count is supported by attention and view masking; the
first-view reference still distinguishes one input.

\subsection{Global and Spatial Camera Estimation}
\label{sec:camera}
The global camera configuration predicts motion from pooled view
features processed by a view-set transformer. Each view is paired with
the reference feature. For each nonreference view, the head predicts a
twist $\xi_k=(\rho_k,\omega_k)\in\mathbb R^6$, with translational and
rotational components $\rho_k,\omega_k\in\mathbb R^3$. The initial pose is
\begin{equation}
 \hat T_k^{(0)}=\operatorname{Exp}_{SE(3)}(\xi_k)=\begin{bmatrix}
 \exp([\omega_k]_\times)&J_l(\omega_k)\rho_k\\0&1
 \end{bmatrix},
\end{equation}
where $[\omega]_\times$ is the skew-symmetric matrix representing the
cross product with $\omega$, and $J_l$ is the $SO(3)$ left Jacobian.
Thus $\operatorname{Exp}_{SE(3)}$ maps a six-dimensional twist to a
rigid transform; its translation is $J_l(\omega)\rho$.
Initial twists have no hard
component bounds. Focal lengths use a positive softplus
parameterization; principal points are unconstrained affine offsets
from the image center. Intrinsic parameters are estimated from RGB
images at inference.

The spatial camera configuration preserves a $16\times16$ grid of
128-dimensional projected features. Reference-to-view cosine
similarities define soft correspondences. For reference location $u$,
\begin{equation}
 a_{uv}=\operatorname{softmax}_v
 \left(\frac{\langle\bar f_1(u),\bar f_k(v)\rangle}{0.1}\right),
 \qquad \tilde v(u)=\sum_v a_{uv}v.
\end{equation}
Here $\bar f$ is unit-normalized. The camera head combines paired
feature content with expected displacement, forward--backward cycle
displacement and normalized matching entropy. A small transformer
summarizes content, while an additional multilayer perceptron (MLP) retains fixed spatial
locations. Reference-visible correspondences derived from training
geometry supervise the matching distribution. This supervision is
absent at inference.

After scene aggregation, an MLP with a zero-initialized output layer
predicts a residual twist $\Delta\xi_k$. Refinement right-multiplies
the initial pose:
\begin{equation}
 \hat T_k=\hat T_k^{(0)}\operatorname{Exp}_{SE(3)}(\Delta\xi_k).
\end{equation}
Each translational and rotational twist component is bounded to 0.08
coordinate units and 0.04 radians, respectively. These are component
bounds in the tangent space, not bounds on the full transform's
translation or rotation magnitude. The reference camera
remains identity. Refinement is learned, rather than a test-time
geometric optimizer.

\subsection{Epipolar Transport and Dense Geometry}
For each target pixel, 24 logarithmically spaced candidate depths define
3D points along its ray. Near and far endpoints are estimated from
training depth. Predicted cameras project each point into the source
views, whose features are sampled bilinearly. Cosine similarities with
temperature 0.15 weight valid samples, producing transported features
and a low-resolution depth prior. This module is inactive during the
initial frozen-backbone training stage and active subsequently and at
evaluation.

The decoder combines transported features with the local RGB pyramid.
Depth is sigmoid-parameterized within the dataset bounds. A bounded
residual augments projective geometry:
\begin{align}
 \hat x_k^c(p)&=\hat D_k(p)\hat A_k^{-1}p+\Delta\hat x_k(p),\\
 \hat x_k^w(p)&=\hat R_k\hat x_k^c(p)+\hat t_k.
\end{align}
Here $p=(u,v,1)^\top$ is a homogeneous pixel coordinate, and
$\hat R_k,\hat t_k$ are the rotation and translation blocks of $\hat T_k$.
Superscripts $c$ and $w$ denote the view's camera frame and the common
first-view frame, respectively.
Each residual coordinate is bounded to 0.1 and regularized. Consequently,
a residual-corrected pointmap need not coincide exactly with the surface
obtained from depth and intrinsics alone. Additional heads predict
confidence, unit normals, support and 64-dimensional correspondence
descriptors. Support denotes CO3D object foreground or TUM valid depth.

\subsection{Observed-View Rendering and Feature Distillation}
Rendering supervision uses a depth-derived Gaussian representation:
predicted depth and cameras provide centers, source RGB provides color,
normals provide orientation, and pixel footprints provide scale.
Confidence controls opacity. Rendering into a second observed view
supplies RGB and depth residuals through the differentiable gsplat
renderer \citep{ye2025gsplat}. This rendering objective is distinct from evaluation on an unseen
view and from residual-corrected pointmap reconstruction.

The geometry-conditioned latent-view prediction (G-LVP) head maps a pooled source
feature and a 20-dimensional condition to a target embedding. The
condition concatenates relative rotation, translation and both source
and target intrinsics. This auxiliary loss does not propagate gradients
through the camera estimates. A frozen
copy of the initial encoder supplies target features. The predictor uses
source features before cross-view scene aggregation, with teacher
embeddings computed from other observed input views.

%% file: chapters/learning_objectives.tex
\section{Learning Objectives and Model Configurations}
\label{sec:objectives}

\subsection{Shared Geometric Supervision}
The model receives complementary local, camera, shape, cross-view,
rendering and auxiliary losses, combined as
\begin{equation}
 \mathcal L=\sum_{q\in\mathcal Q}\lambda_q\mathcal L_q,
\end{equation}
where $\mathcal Q$ indexes the individual terms and $\lambda_q$ includes
the configured coefficient and training-stage multiplier.
Local depth uses Smooth L1 and a log-depth residual term
$\mathbb E[e^2]-0.85\mathbb E[e]^2$, where
$e=\log\hat d-\log d$. Motivated by scale-invariant log-depth
supervision \citep{eigen2014depth}, our coefficient 0.85 retains
sensitivity to mean log-depth error: the term equals
$\operatorname{Var}(e)+0.15\mathbb E[e]^2$.
Normals use orientation supervision, and edge-aware smoothness
regularizes depth. Camera supervision compares first-view-relative
translations and rotations, normalized intrinsics and pairwise relative
poses. Pointmaps receive camera-frame supervision and a robust
similarity-aligned world-point loss. The similarity fit is held fixed
for differentiation, and its valid reference support is independent of
predicted confidence.

Cross-view terms measure world-point consistency, image-coordinate
cycles, photometric reprojection and depth reprojection, subject to
reference visibility. A projection-coverage penalty discourages
solutions that move points outside the other image. Geometry-derived
correspondences supervise dense descriptors through attraction and
contrastive terms. The support head uses binary cross-entropy and Dice
loss. Rendering compares RGB and depth in a second observed view.
Weights and mathematical details appear in
Appendix~\ref{app:objectives}.

\subsection{Relative Objectives and Calibration}
\label{sec:relative-objectives}
We additionally test a joint configuration, called \emph{relative
objectives}, on top of spatial camera estimation. It combines robust
relative and log-depth losses with a reduced absolute-depth weight;
translation supervision normalized by the reference camera-baseline
distance within each training window; direct supervision
of initial poses; and a penalty when refinement raises the supervised
pose loss. Calibration has a dedicated view-set context and a larger
loss weight. Its loss compares log focal lengths and normalized
principal points. Predicted intrinsics still enter all geometric
computations, but gradients from downstream geometry are stopped at
their output. Calibration-specific parameters therefore receive direct
calibration supervision without downstream geometric gradients through
the predicted intrinsics. Shared encoder features still receive losses
from several tasks, so the tasks remain coupled.

The configuration also changes confidence and distillation objectives.
These interventions are evaluated jointly. Differences from the spatial
camera configuration therefore cannot be attributed to one objective,
calibration isolation or architecture component in isolation.

\subsection{Confidence and Latent Targets}
For the global and spatial camera configurations, confidence is fitted
to $c^*=\exp(-|\hat d-d|/\tau)$ on valid pixels, with zero targets on
invalid pixels. Here $\tau=0.08$ in CO3D coordinates and $0.10$\,m on
TUM. This is an absolute-error score, not a probability of a specified
relative-accuracy event. The relative-objective configuration instead
uses the binary target
\begin{equation}
 y_p=\mathbf 1\!\left[\frac{|\hat d_p-d_p|}{d_p}\leq0.10\right]
\end{equation}
and binary cross-entropy, with predicted depth held constant when
differentiating the target. Reliability is evaluated for this stated event.
Raw confidence values across the two target definitions do not have the
same interpretation.

Let $\hat z_{ij}$ denote the target embedding predicted from source view
$i$ and its camera condition for target view $j$, and let $z_j$ denote
the frozen teacher's pooled embedding of view $j$.
The basic distillation loss averages $1-\cos(\hat z_{ij},z_j)$ over
ordered pairs $i\ne j$. The relative-objective configuration standardizes
teacher features using training-frame statistics, normalizes them, and
adds an InfoNCE-style within-window target retrieval loss
\citep{oord2018cpc}. With both predicted and transformed teacher
embeddings normalized to unit length, the objective is
\begin{align}
 \mathcal L_{\rm distill}
 &=\mathbb E_{i\ne j}[1-\hat z_{ij}^{\top}z_j]+0.25\mathcal L_{\rm retrieval},\\
 \mathcal L_{\rm retrieval}
 &=\mathbb E_{i\ne j}\left[-\log
 \frac{\exp(\hat z_{ij}^{\top}z_j/0.1)}
 {\sum_{l=1}^{K}\exp(\hat z_{ij}^{\top}z_l/0.1)}\right].
\end{align}
The denominator contains all $K$ observed teacher views as candidate
targets. Validation images supply teacher targets but do not fit the
standardization statistics. Test images enter neither reconstruction
training nor these statistics. Correct-versus-shuffled cosine and
retrieval measure target selectivity. Neither alone measures
a causal improvement in reconstruction.

%% file: chapters/experimental_protocol.tex
\section{Experimental Protocol}
\label{sec:protocol}

\subsection{Datasets and Sequence Separation}
\paragraph{CO3D.}
We use twelve sequences from six object categories in CO3D
\citep{reizenstein2021co3d}. Eight sequences provide 425 training windows,
one bench sequence provides 56 validation windows, and two apple
sequences plus one cake sequence provide 168 test windows. Sequences
are disjoint across partitions. Each temporal window contains ten
frames sampled at frame stride four and window step three. Depth PNGs
are interpreted as float16 bit patterns and multiplied by their
annotation scale. Camera poses are converted from the dataset convention
to the camera convention used for unprojection. Bounds are
$[0.03,50]$ in CO3D coordinates, which are not assumed to be meters.
The test coverage is limited to apple and cake; the experiments do not
measure generalization over all CO3D categories.

\paragraph{TUM RGB-D.}
The indoor model is trained independently on TUM RGB-D
\citep{sturm2012tum}. Four sequences provide 1843 training windows:
fr1/room, fr2/desk, fr2/xyz and fr3/long\_office\_household.
Fr1/desk provides 97 validation windows, while
fr3/structure\_texture\_far provides 163 test windows. Ten-frame windows
use frame stride ten and window step five. RGB and depth are associated
within 20\,ms; camera trajectories are interpolated at RGB timestamps
after consolidating duplicate pose timestamps. PNG depth is divided by
5000 to obtain meters, with valid range $[0.10,6.0]$\,m.

The registered-depth convention uses the ROS-default intrinsics
$(525,525,319.5,239.5)$, following TUM's guidance when avoiding
undistortion of registered depth \citep{tumformats}.
Images are scaled from $640\times480$ to $512\times384$ and padded
vertically to $512\times512$. The corresponding intrinsic parameters
are $(420,420,255.6,255.6)$ in every sequence. These fixed labels make
TUM a test of adherence to the registered calibration convention,
rather than recovery of varying physical camera calibration.
The 163 test windows overlap and contain only 181 unique RGB frames.

\subsection{Configurations and Optimization}
We compare three CO3D configurations: \emph{global camera},
\emph{spatial camera}, and \emph{relative objectives}.
The first two differ in camera reasoning and correspondence supervision;
the third uses the joint changes in Section~\ref{sec:relative-objectives}.
The global camera configuration is also trained independently on TUM.
There is no TUM evaluation of the other two configurations.

All primary experiments use $512\times512$ images and 20 epochs.
The first five train the prediction modules with the backbone frozen;
the next ten adapt its last transformer block and normalization;
the final five freeze the backbone and increase alignment weights.
AdamW \citep{loshchilov2019adamw} uses learning rates
$1.5\times10^{-4}$ for prediction modules
and $2\times10^{-6}$ for the backbone, weight decay $10^{-4}$ and
gradient clipping at norm 1. Training uses mixed precision on two
16-GB NVIDIA Tesla T4 GPUs. One sample per GPU and accumulation over
two steps give an effective batch size of four. Primary evaluation
uses 32-bit floating-point arithmetic.
Training seed is 42 and evaluation sampling seed is 123.

Stages one and three sample $K\in\{3,5,8\}$ with probabilities
$(0.45,0.35,0.20)$; stage two samples $K\in\{3,5\}$ with probabilities
$(0.65,0.35)$. View dropout has probability 0.15 while retaining at
least three views. Evaluation deterministically selects five evenly
spaced views from every ten-frame window. Appendix~\ref{app:data}
provides the sequence split and training-depth statistics.

\subsection{Model and Threshold Selection}
Global-camera models are selected by minimizing mean validation AbsRel.
Spatial-camera and relative-objective models are selected by minimizing
\begin{equation}
\label{eq:selection}
 S=\overline{\mathrm{AbsRel}}+
 0.5\,\overline{\mathrm{ATE}/\mathrm{ATE}_{\rm id}}+
 0.25\,\overline{\mathrm{RPE}_{R}}/180,
\end{equation}
where rotation is in degrees. The identity ratio is computed per window
with denominator lower bound $\max(0.001d_{\rm near},10^{-6})$,
where $d_{\rm near}$ is the training-depth first percentile. Selected epochs
are 14, 12 and 9 for the three CO3D configurations, respectively, and
18 for TUM. All selected CO3D models therefore precede the final
alignment-focused stage, while the selected TUM model is from that stage.
Consequently, the CO3D comparisons include a change in
selection criterion as well as architecture and supervision.

For each selected model, the confidence threshold is chosen to minimize
mean raw Chamfer distance on the validation set, with the predicted
support threshold fixed at 0.5. The selected confidence threshold is
0.005 for all models. Ties select the lowest threshold. These operating points are
selected using validation data alone within each configuration;
test-informed configuration development is discussed in
Section~\ref{sec:discussion}. Supplementary component studies have their
own protocols, including post-hoc surface thresholds in one suite;
these are stated separately in Section~\ref{sec:components} and
Appendix~\ref{app:additional}.

\subsection{Measurements and Simple Baselines}
We report depth root mean squared error (RMSE), absolute relative error
(AbsRel) and ratio-threshold accuracies $\delta_n$; first-view-anchored
camera absolute trajectory error (ATE); adjacent-view translation and
rotation relative pose error (RPE); normalized intrinsics mean absolute
error (MAE); and camera-frame pointmap RMSE. ATE and RPE follow the
trajectory-error terminology of the TUM benchmark \citep{sturm2012tum},
with the alignment conventions specified here.
Primary ATE fits no further alignment. Each metric is computed per
window and then averaged. A separate ten-view TUM analysis fits
rotation and translation within each window, without scale; it is
reported separately from the primary five-view metric.

Surface evaluation uses a fixed reference set from valid depth and
object support on CO3D, or valid depth on TUM. Predicted confidence
and support filter only the predicted set. Nearest-neighbor distances
are computed within each window, followed by an average over windows.
We report unsquared Chamfer distance, the sum of accuracy and
completeness, and F-score at distances 0.05, 0.10 and 0.20. These
thresholds are meters on TUM and dataset units on CO3D. Similarity
alignment is fitted separately for each window before filtering;
raw and aligned results are reported together. Empty predictions receive
zero F-score and a fixed distance penalty, rather than being omitted.
Complete definitions appear in Appendix~\ref{app:metrics}.

Identity cameras place all centers at the reference origin. Constant
depth uses the median valid training depth, and constant calibration
uses the training-median normalized intrinsics. The controls use the
same evaluation support as the learned predictions. They are elementary
priors, not competitors with state-of-the-art reconstruction methods.
A comparison against such methods or alternative pretrained encoders
under a common protocol is outside the present experiments.

%% file: chapters/results.tex
\section{Results}
\label{sec:results}
\input{chapters/co3d_results}
\input{chapters/tum_results}
\input{chapters/reconstruction_evidence}

\subsection{Confidence Reliability and Latent Discrimination}
\label{sec:confidence-distillation}
\begin{figure*}[t]
\centering
\includegraphics[width=0.95\textwidth]{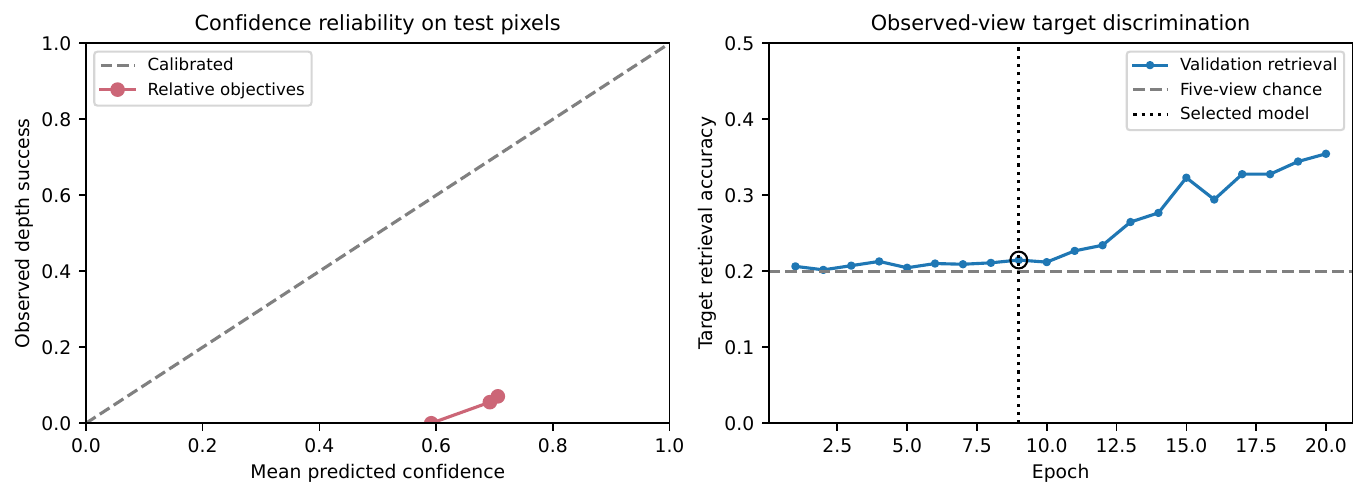}
\caption{Confidence reliability and target retrieval for the relative-objective CO3D model.
Left: observed relative-depth success versus predicted confidence on
valid test pixels. Right: validation target retrieval over training,
with the selected model marked. Later retrieval improvements are not
measurements of the model used for the primary reconstruction tables.}
\label{fig:auxiliary-objectives}
\end{figure*}

Confidence and feature similarity test different aspects of the
auxiliary objectives. For the relative-objective CO3D model,
confidence denotes probability of depth relative error at most 10\%.
Across valid test pixels its mean is 0.6971, while observed success is
0.0610. Ten-bin pixel-weighted expected calibration error (ECE)
\citep{guo2017calibration} is 0.6360. Averaging metrics over windows
instead gives ECE 0.6310, binary Brier score 0.4587
\citep{brier1950verification} and success fraction 0.0705. The two aggregations differ because
windows contain unequal valid-pixel counts, but both show substantial
overconfidence. A constant probability of 0.5 gives Brier 0.25, and
zero gives mean Brier 0.0705 on the same event labels. These are
confidence controls conditional on the model's depth errors, not
alternative depth estimators. Figure~\ref{fig:auxiliary-objectives}
shows the reliability and selected-model retrieval measurements.

Threshold sensitivity also fails to support useful filtering. Spatial
camera surface scores are unchanged at thresholds 0.005--0.05;
at 0.2, 163 of 168 predicted clouds are empty. Relative-objective
masked scores are unchanged through 0.4, and all predictions are empty
at 0.8. On TUM, increasing threshold from 0.005 to 0.05 worsens raw
Chamfer from 0.7515 to 0.8076\,m and F@20\,cm from 0.1191 to 0.1035.
Figure~\ref{fig:confidence-filtering} shows the threshold curves together
with the fraction of empty predicted clouds for both datasets.
Confidence scores are therefore meaningful only in relation to their
stated targets and observed calibration.

The basic observed-view distillation objective attains high cosine
similarity with weak target specificity. At the selected spatial camera
epoch, training correct-target cosine is 0.992164 and shuffled-target cosine is
0.992270; shuffled targets score higher in 17 of 20 training epochs.
The TUM selected epoch has a small positive training difference,
0.997951 versus 0.996445. Standardized targets and contrastive retrieval
produce validation retrieval 0.2148 for the selected relative-objective
model, against five-view chance 0.20. Retrieval reaches 0.3546 at the
end of training, but that later model supplies none of the primary
reported reconstruction scores. Absolute cosines across standardized
and unstandardized spaces are not directly comparable. These results
show why high latent agreement alone is insufficient evidence of
view-specific prediction or better reconstruction.

\subsection{Three-View Component Comparisons}
\label{sec:components}
The three-view component studies provide complementary, protocol-specific
evidence. A five-configuration CO3D distillation suite used a 3/6/3-epoch
schedule and 195 evaluation windows. Adding the conditioned objective
at weight 0.025 increased depth-target RMSE from 0.5153 to 0.6339 while
leaving ATE nearly unchanged (1.7438 versus 1.7359). Its minimum
aligned Chamfer across the confidence thresholds evaluated on the test
set was lower, 0.1974 versus 0.3171. However, those surface
scores use test-selected thresholds, prediction-dependent reference
support and pooled local windows. Their depth targets use an
image-value rescaling heuristic rather than the primary annotation
decoding, as detailed in Appendix~\ref{app:additional}.
They are exploratory comparisons,
not estimates under the primary evaluation protocol or evidence for a
causal benefit in the five-view model.

In a separate short-schedule TUM component suite, freezing the backbone
lowered RMSE from 0.8570 to 0.7159\,m but increased ATE from 0.0811 to
0.0943\,m. Reinitializing the common learnable backbone layers gave
RMSE 0.7051\,m and ATE 0.0942\,m. Removing epipolar fusion lowered
ATE to 0.0501\,m while increasing RMSE to 0.8955\,m; removing
confidence supervision and weighting lowered ATE to 0.0440\,m while
increasing RMSE to 0.8924\,m. The suite supports a depth--pose trade-off
within that setting. It neither isolates the contribution of video
pretraining to the primary model nor establishes that every module is
beneficial. Full component results and the distinct random-window,
noisy-pose experiment are in Appendix~\ref{app:additional};
Figure~\ref{fig:component-tradeoffs} visualizes the competing depth and
camera effects within each component suite.

%% file: chapters/co3d_results.tex
\subsection{CO3D Reconstruction}
\label{sec:co3d-results}

\begin{table*}[t]
\centering\small
\caption{CO3D results on the same 168 five-view windows. Depth and
translation follow dataset coordinates; they are not interpreted as
meters. Intrinsics error uses normalized parameters. Each learned row
is one training seed. Controls use training-median depth, identity
cameras and training-median intrinsics, respectively.}
\label{tab:co3d-dense}
\begin{tabular}{lrrrrrr}
\toprule
Configuration & RMSE $\downarrow$ & AbsRel $\downarrow$ & ATE $\downarrow$ & RPE rot. $\downarrow$ & $\delta_1$ $\uparrow$ & Intrinsics MAE $\downarrow$\\
\midrule
Global camera & 2.8296 & 0.2746 & 9.4225 & $30.4288^\circ$ & 0.3921 & 0.0540\\
Spatial camera & 2.8821 & 0.2900 & 6.6586 & $22.6679^\circ$ & 0.4760 & 0.1133\\
Relative objectives & 3.1874 & 0.3171 & 6.8492 & $20.4223^\circ$ & 0.2967 & 0.0970\\
Simple controls & 2.9483 & 0.2902 & 6.0516 & --- & --- & 0.1296\\
\bottomrule
\end{tabular}
\end{table*}

\paragraph{The spatial-camera configuration has lower motion error.}
Table~\ref{tab:co3d-dense} shows a 29.33\% mean ATE reduction from
global to spatial camera estimation. ATE improves in all 168 paired
windows, and rotation RPE falls from $30.43^\circ$ to $22.67^\circ$.
Nevertheless, spatial camera ATE remains 10.03\% above the identity
baseline. On the two apple sequences, ATE is 6.3767 and 11.1314 versus
identity values 4.7718 and 7.5386; on cake it is 2.4675 versus 5.8443.
The aggregate improvement therefore coexists with strongly uneven
sequence-level camera accuracy. Spatial camera refinement slightly
increases ATE, from 6.6350 to 6.6586.

\paragraph{Depth and calibration do not improve with camera ATE.}
Mean AbsRel increases by 5.59\%, and normalized intrinsics MAE more
than doubles. Although depth RMSE and AbsRel improve in 112 of the
168 paired windows, a large regression on one apple sequence worsens
their overall means. Spatial camera AbsRel is only 0.10\% below the
constant-depth control. Its better $\delta_1$ also accompanies a lower
$\delta_2$ (0.9012 versus 0.9672). Improvements at one ratio threshold
therefore do not imply a uniformly better depth-error distribution.

\paragraph{Relative objectives favor some outputs and degrade others.}
The joint relative-objective configuration reduces intrinsics MAE by
14.37\% and rotation RPE by 9.91\% relative to spatial camera estimation.
It increases depth RMSE by 10.60\%, AbsRel by 9.36\% and camera ATE
by 2.86\%. Its RMSE and AbsRel are 8.11\% and 9.25\% worse than
constant depth, and ATE is 13.18\% worse than identity. The lower
validation selection score (0.4119 versus 0.4505) does not translate
into lower test depth or camera error. Refinement still raises ATE,
from 6.8203 to 6.8492, despite the additional supervised penalty.

\begin{table*}[t]
\centering\small
\caption{CO3D surface reconstruction with fixed reference support and
validation-selected confidence 0.005. Chamfer uses dataset units;
F@0.20 uses the same scale. Similarity fitting is performed separately
for each window and uses reference correspondences.}
\label{tab:co3d-surfaces}
\begin{tabular}{lrrrr}
\toprule
 & \multicolumn{2}{c}{Raw coordinates} & \multicolumn{2}{c}{Fitted similarity}\\
Configuration & Chamfer $\downarrow$ & F@0.20 $\uparrow$ & Chamfer $\downarrow$ & F@0.20 $\uparrow$\\
\midrule
Global camera & 3.7868 & 0.0368 & 0.5987 & 0.4047\\
Spatial camera & 3.5535 & 0.0333 & 0.6447 & 0.3943\\
Relative objectives & 4.4374 & 0.0265 & 0.5573 & 0.4476\\
\bottomrule
\end{tabular}
\end{table*}

\begin{figure*}[t]
\centering
\includegraphics[width=\textwidth]{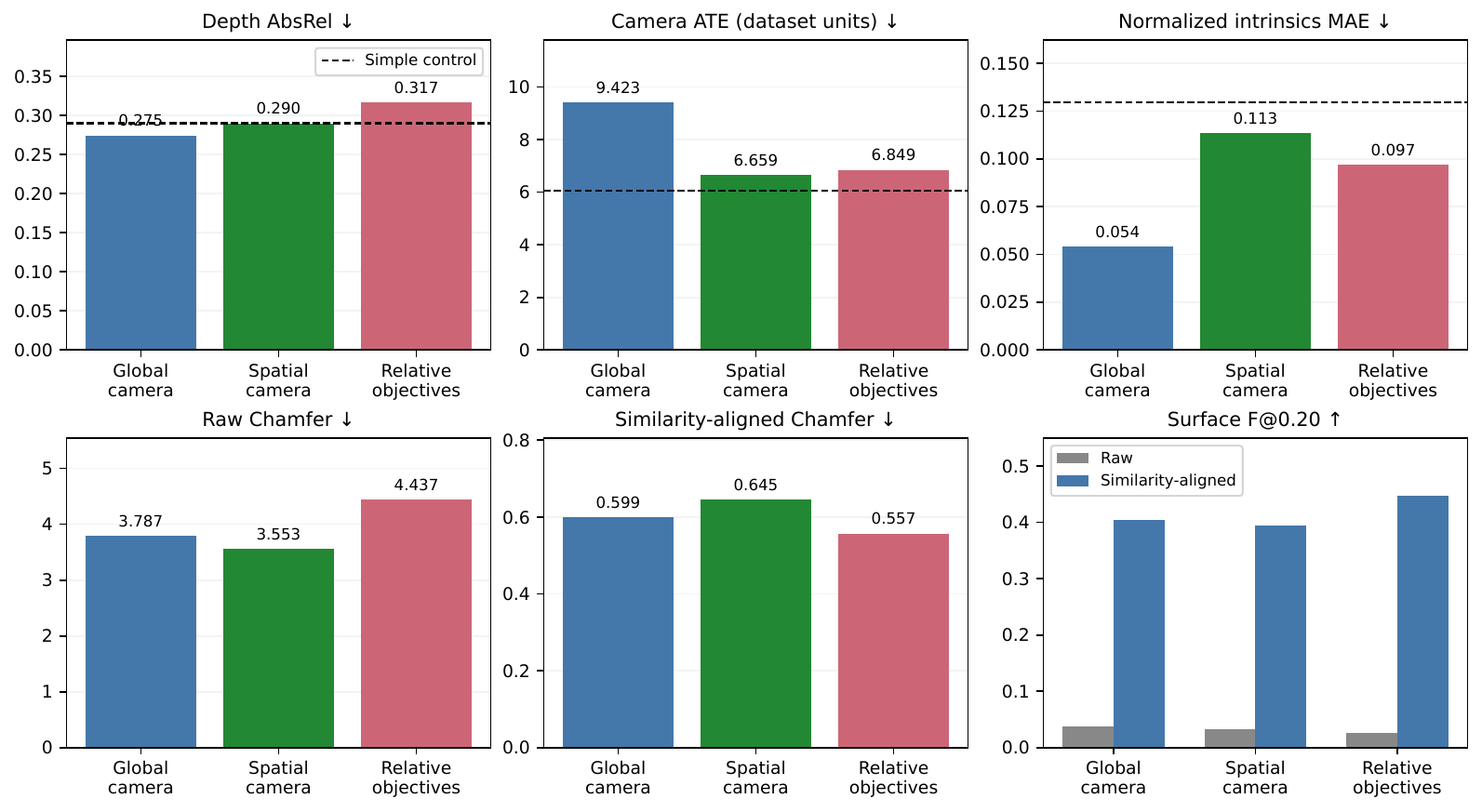}
\caption{CO3D geometric trade-offs. The spatial-camera configuration has
lower camera ATE than the global-camera configuration. The
relative-objective configuration has lower intrinsics and aligned-surface
errors than the spatial-camera configuration, but higher depth and
raw-surface errors.
Dashed controls are training-median depth, identity cameras and
training-median intrinsics. The configurations differ jointly in
architecture, supervision and selection criterion.}
\label{fig:geometry-tradeoffs}
\end{figure*}

\paragraph{Aligned shape and raw reconstruction give different conclusions.}
Table~\ref{tab:co3d-surfaces} separates surface accuracy before and
after reference fitting. Relative objectives lower aligned Chamfer from
0.6447 to 0.5573 and increase aligned F@0.20 from 0.3943 to 0.4476.
Yet raw Chamfer increases by 24.88\%, and raw F@0.20 falls from 0.0333
to 0.0265. This is a shape improvement conditional on a reference-derived
transformation, rather than improved unaligned reconstruction in the
first-view frame.
Figure~\ref{fig:geometry-tradeoffs} summarizes these competing trends.
Figure~\ref{fig:co3d-depth-predictions} compares the full depth
predictions on identical RGB views, and
Figure~\ref{fig:training-generalization} shows their validation
behavior over the training schedule.

%% file: chapters/tum_results.tex
\subsection{Indoor Reconstruction}
\label{sec:tum-results}

\begin{table}[t]
\centering\small
\caption{TUM primary results on 163 five-view windows. Depth and
translation are in meters. The model uses global camera estimation
and is trained on TUM independently of CO3D. Controls use training-median
depth, identity cameras and training-median calibration.}
\label{tab:tum-controls}
\begin{tabular}{lrr}
\toprule
Metric & Model & Control\\
\midrule
Depth RMSE $\downarrow$ & 1.0159 & 0.8822\\
AbsRel $\downarrow$ & 0.4279 & 0.2516\\
Raw window ATE $\downarrow$ & 0.3063 & 0.3694\\
Normalized intrinsics MAE $\downarrow$ & 0.0617 & 0.0000\\
\bottomrule
\end{tabular}
\end{table}

\paragraph{Camera and depth errors diverge.}
Table~\ref{tab:tum-controls} compares the model with the simple controls.
The indoor model reduces ATE by 17.08\% relative to identity cameras,
improving in 144 of 163 windows. However, mean depth RMSE is 15.16\%
worse than constant depth, and mean AbsRel is 70.07\% worse. AbsRel
exceeds the depth control in every window. Depth $\delta_1$ is 0.0994,
translation RPE is 0.1554\,m, rotation RPE is $3.7418^\circ$ and
camera-frame pointmap RMSE is 0.6288\,m.

\begin{figure*}[t]
\centering
\includegraphics[width=0.98\textwidth]{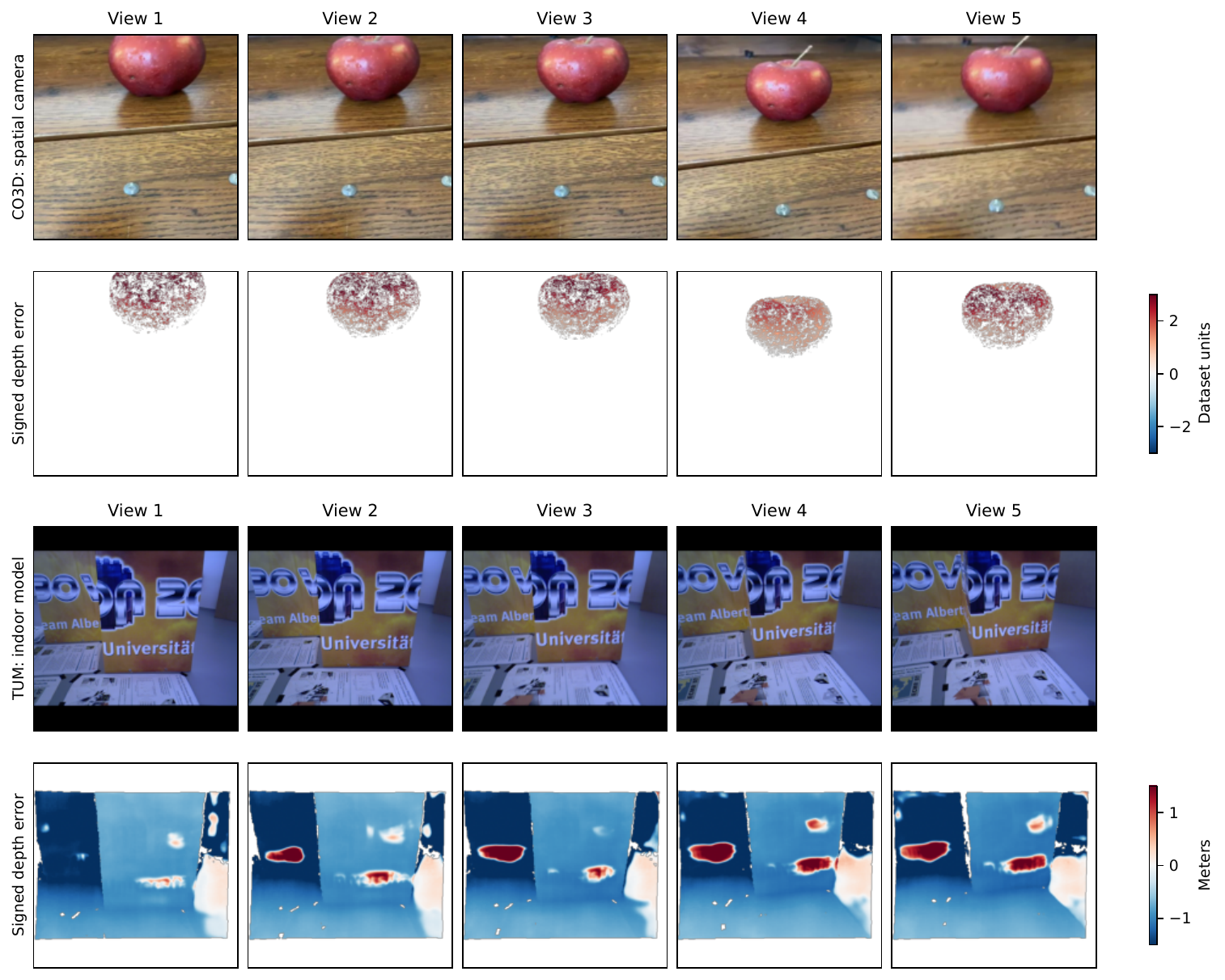}
\caption{First test window for each setting:
CO3D spatial camera and the independently trained TUM model. Each RGB
row is followed by signed depth error, predicted minus reference.
White pixels lack valid reference depth. Color ranges are $[-3,3]$ in
CO3D coordinates and $[-1.5,1.5]$\,m on TUM, with saturation beyond
the limits. Blue indicates depth underestimation. These windows
illustrate failure patterns rather than the distribution of all errors.}
\label{fig:qualitative-depth}
\end{figure*}

Validation AbsRel is 0.1411, substantially below test AbsRel 0.4279.
Since these measurements come from different sequences, the gap concerns
sequence generalization rather than training-set fit alone. The three
qualitative windows show mean predicted depths 1.18--1.32\,m versus
reference means 2.16--2.19\,m. Camera-center dispersion is measured as
the root mean square (RMS) Euclidean distance of the nonreference centers from their
centroid within each window. Predicted dispersion is 0.007--0.014\,m
versus reference values of 0.174--0.232\,m.
These examples exhibit depth underestimation and nearly coincident
nonreference centers, even though mean ATE is below the identity control.
The examples do not establish the same camera-spread pattern in every
window. Figure~\ref{fig:qualitative-depth} shows signed depth errors
for the first test window in each setting.

\paragraph{Refinement and trajectory alignment.}
Refinement worsens primary test ATE by 14.18\%, from 0.2682 to
0.3063\,m, while slightly improving validation ATE from 0.6958 to
0.6902\,m. Refinement therefore has different effects on the validation
and test sequences. Selecting an unrefined model after observing this test
result would constitute a new, test-informed choice.

The separate ten-view analysis gives rigid-aligned ATE 0.1704\,m,
compared with 0.1956\,m for identity positions under the same alignment.
Translation and rotation RPE are 0.0796\,m and $2.3978^\circ$.
This lower ATE cannot be attributed to additional views alone, because
alignment also differs from the primary metric. Both analyses concern
overlapping local windows, not a stitched sequence trajectory or SLAM
system with loop closure.

\paragraph{Calibration and surface reconstruction.}
Constant registered-calibration labels make the training-median
intrinsics control exact. A learned MAE of 0.0617 therefore measures
deviation from a fixed convention, not successful recovery of diverse
camera calibration.

\begin{figure*}[t]
\centering
\includegraphics[width=0.96\textwidth]{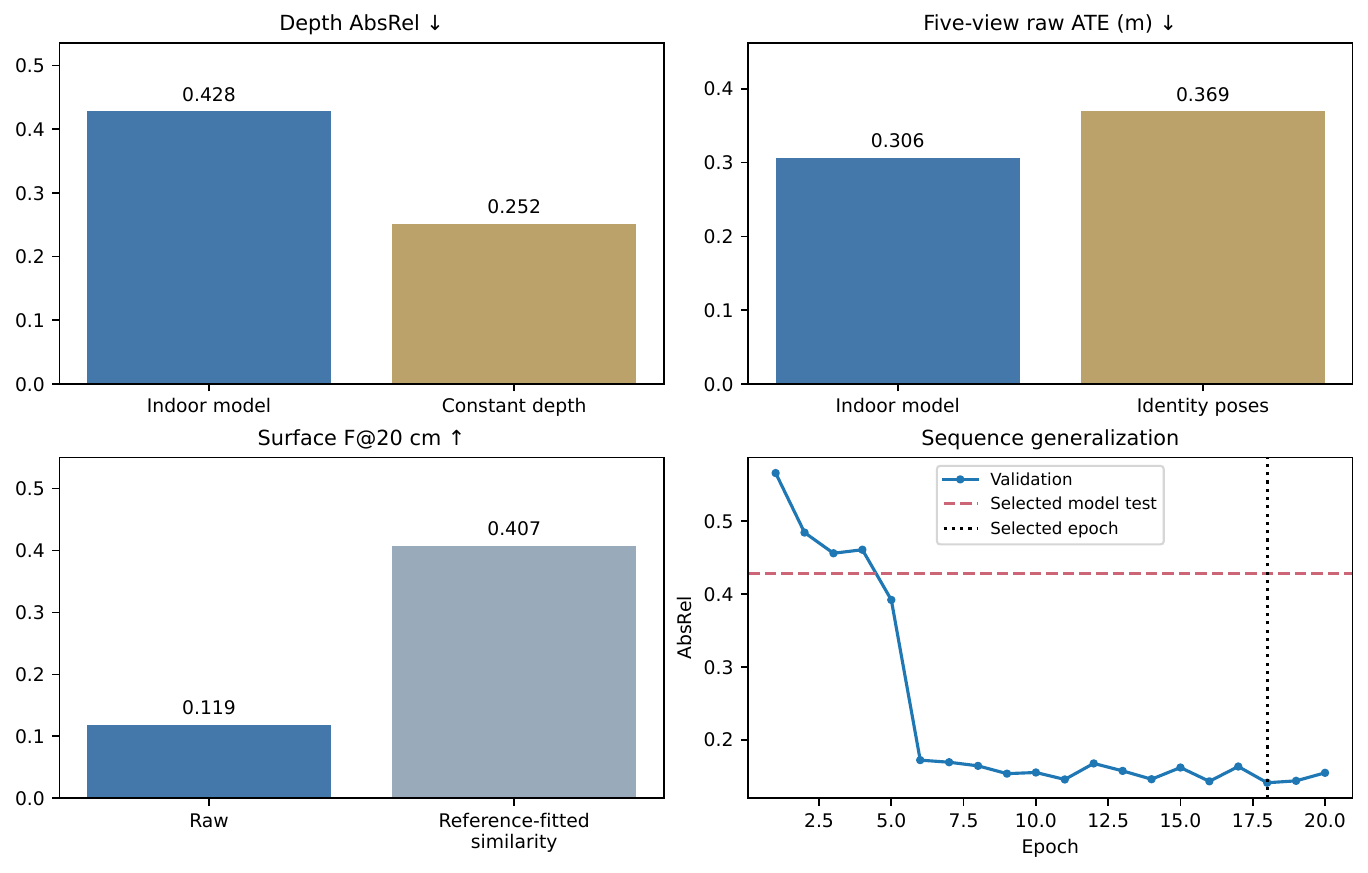}
\caption{Indoor generalization and geometry. The model improves camera
ATE over identity but has higher depth error than training-median depth. Reference fitting
substantially changes surface F-score. Validation AbsRel improves during
training while the selected model's held-out-sequence error remains
higher. The test line represents one selected model, not repeated test
evaluation over training.}
\label{fig:indoor-generalization}
\end{figure*}

At confidence 0.005, raw Chamfer is 0.7515\,m, with F@5\,cm 0.0302,
F@10\,cm 0.0600 and F@20\,cm 0.1191. Reference-fitted similarity gives
Chamfer 0.5192\,m and F@20\,cm 0.4075, with mean fitted scale 0.6893.
Mean raw precision and recall at 20\,cm are 0.0677 and 0.5622;
F-score is averaged after each window's harmonic mean. Observed-view
rendering has coverage 0.9850, RGB MAE 0.1627 and depth MAE 0.9743\,m.
High image coverage therefore coexists with substantial geometric error.
Figure~\ref{fig:indoor-generalization} relates depth, camera and surface
measurements to validation behavior.
The complete five-view depth comparison appears in
Figure~\ref{fig:tum-depth-predictions}. Table~\ref{tab:tum-surfaces}
reports the surface measurements before and after reference fitting.

\begin{table}[t]
\centering\small
\caption{TUM surface reconstruction at confidence 0.005. Chamfer is in
meters; F-scores use 5, 10 and 20\,cm. The same fixed reference support
is used before and after per-window similarity fitting.}
\label{tab:tum-surfaces}
\begin{tabular}{lrrrr}
\toprule
Alignment & Chamfer & F@5 & F@10 & F@20\\
\midrule
Raw & 0.7515 & 0.0302 & 0.0600 & 0.1191\\
Similarity & 0.5192 & 0.1201 & 0.2267 & 0.4075\\
\bottomrule
\end{tabular}
\end{table}

%% file: chapters/reconstruction_evidence.tex
\subsection{Camera Motion, Surfaces and Observed-View Rendering}
\label{sec:visual-geometry}
\begin{figure*}[t]
\centering
\includegraphics[width=0.95\textwidth]{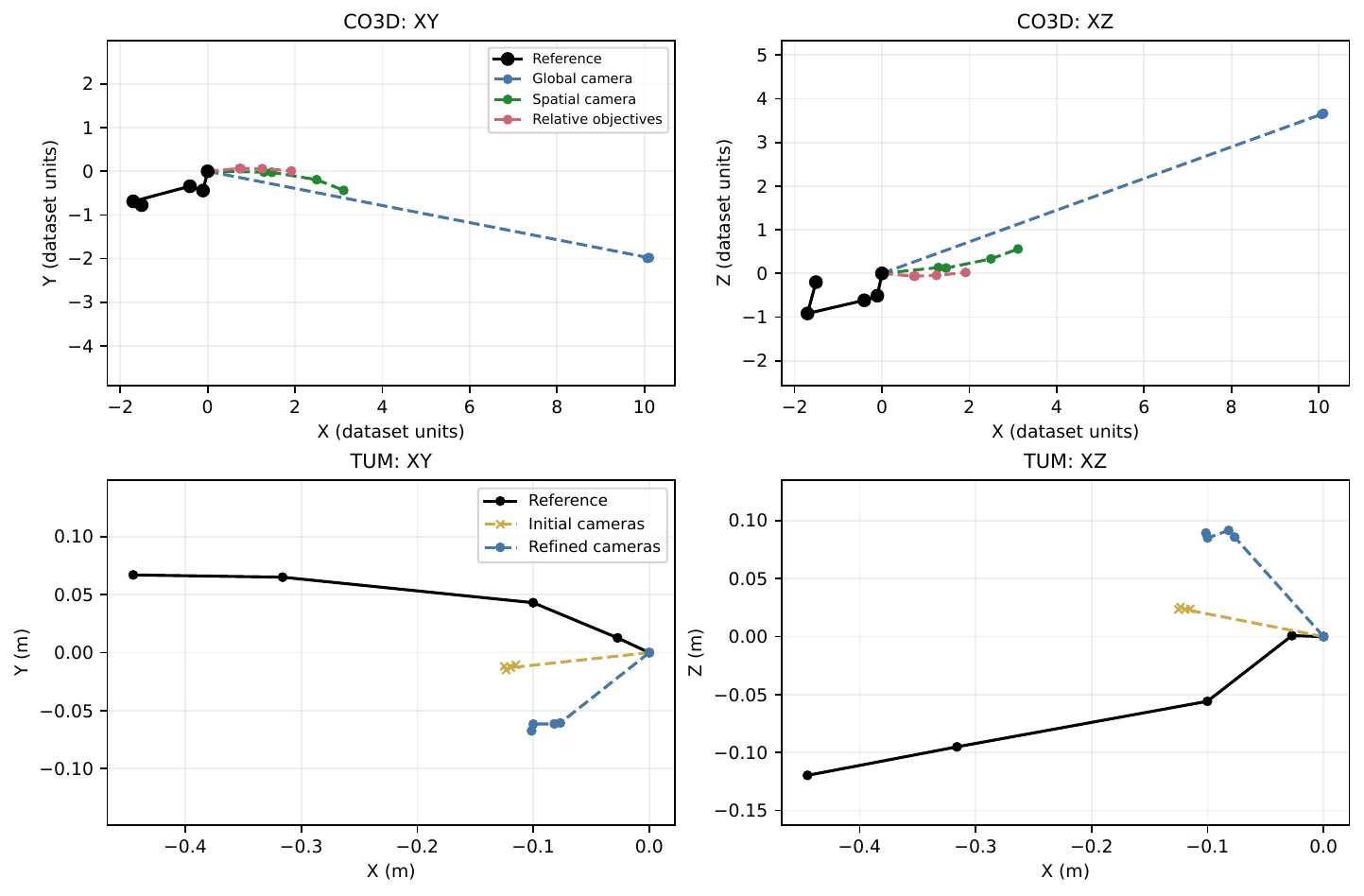}
\caption{Five-view camera centers in the first test window of CO3D and
TUM. Orthographic XY and XZ projections use the same reference frame
for predictions and labels, with no fitted trajectory alignment. Lines
connect the sampled views in temporal order. CO3D compares the three
configurations on identical observations; TUM compares initial and
refined predictions from its independently trained model. These local
windows do not constitute a full-sequence trajectory.}
\label{fig:camera-trajectories}
\end{figure*}

Figure~\ref{fig:camera-trajectories} compares camera centers for the
same five observations in each setting, without fitting a trajectory
alignment. The CO3D configurations predict different motion even when
given identical images. In the illustrated object window, all three
predict lateral motion on the opposite side of the reference origin
from the reference camera path. The aggregate ATE reduction therefore
does not imply agreement with the reference path in every window.
The indoor example has closely spaced
nonreference predictions despite nonzero reference motion. A lower
window ATE than identity therefore need not imply recovery of the
within-window path. These examples illustrate the distinction between
mean camera error and the spatial arrangement of the predicted centers.

\begin{figure*}[p]
\centering
\includegraphics[width=0.94\textwidth,height=6.9in,keepaspectratio]{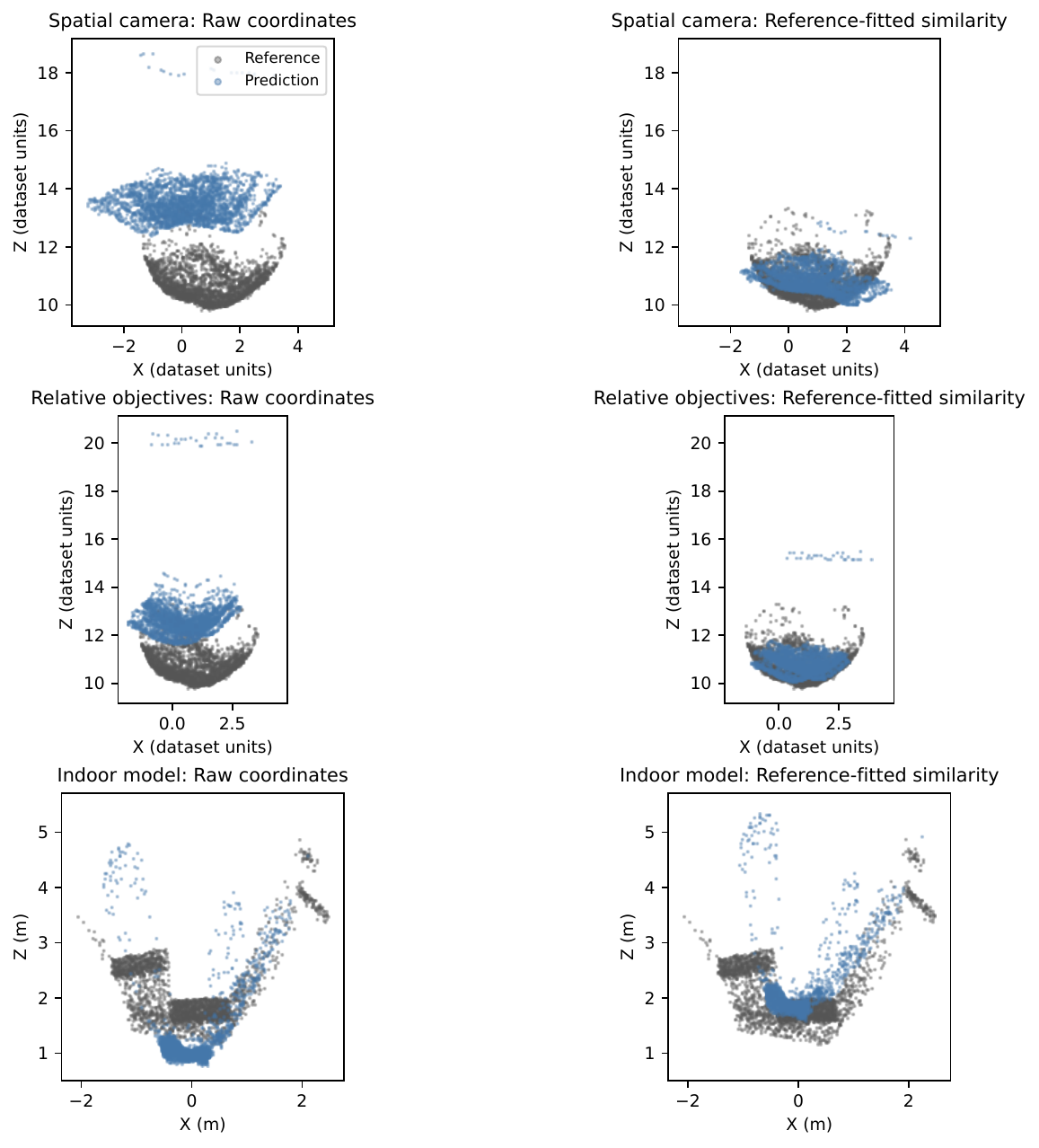}
\caption{XZ projections of reconstructed pointmaps for the first test
window. Gray denotes the fixed reference surface; blue denotes predicted
points after confidence 0.005 and support filtering. The right column
uses a separate reference-fitted similarity transform per window.
Axes are identical within each row, and points are subsampled only for
display. The two CO3D configurations share their reference window; the
indoor row uses TUM meters. Shape after reference fitting is distinct
from unaligned accuracy in the first-view coordinate frame.}
\label{fig:surface-reconstructions}
\end{figure*}

Figure~\ref{fig:surface-reconstructions} shows raw and similarity-aligned
pointmaps against the reference surfaces. Each row uses one temporal
window and fixed axis limits across the two panels. Reference fitting
changes placement and scale, but residual surface discrepancies remain.
The numerical distinction is substantial on TUM: F@20\,cm changes
from 0.1191 raw to 0.4075 aligned. On CO3D, the relative-objective
configuration has better mean aligned shape but worse raw surfaces
than spatial camera estimation. The relative ranking therefore depends
on the alignment convention.

Rendering gives a further separation between appearance and geometry.
From spatial camera estimation to relative objectives, CO3D RGB MAE
falls from 0.2062 to 0.1686 (18.20\%), while rendered-depth MAE rises
from 2.7765 to 3.1039 (11.79\%). Both have image coverage above 99.9\%.
The RGB improvement accompanies worse raw Chamfer, rather than a
consistent improvement in all outputs. The independently trained indoor
model likewise combines 98.50\% image coverage with depth-rendering MAE
0.9743\,m. These measurements concern a second observed view, with
source RGB providing Gaussian color; they do not establish unseen-view
synthesis quality. Table~\ref{tab:observed-rendering} reports all four
models, and Table~\ref{tab:complete-surfaces} gives complete surface
measurements at the selected operating points.

%% file: chapters/discussion.tex
\section{Discussion}
\label{sec:discussion}

\subsection{Representation Quality and Geometric Accuracy}
The results distinguish a usable visual representation from accurate
camera-consistent reconstruction. The spatial-camera configuration has
lower CO3D camera error, but depth, calibration and surfaces do not improve
together. On TUM, lower ATE than the identity baseline coexists with severe depth
underestimation. These are empirical properties of the tested models;
they do not imply that a specific backbone intrinsically lacks geometry.
No primary experiment holds all other factors fixed while replacing
the video-pretrained encoder.

The model offers several ways to reduce one residual without correcting
another. Pointmap residuals can adjust projective depth points;
calibration can affect ray directions; translations and depth jointly
set relative scene scale. A fitted similarity transformation can remove
scale and placement errors before a shape metric is computed.
The raw/aligned gap therefore has a concrete interpretation: the
predicted point configuration has lower measured surface error after
reference-derived alignment than in its predicted coordinate frame.
This does not prove that the
cause is only scale, since rotation and translation are also fitted.
The fitted scale acts on world pointmaps rather than depth alone.
It depends jointly on predicted rays, camera placement and pointmap
residuals. A fitted scale below one can therefore coexist with
underestimated camera-frame depth, as in the indoor experiment;
it should not be interpreted as a standalone measure of depth bias.

\subsection{Joint Objectives Do Not Guarantee Joint Improvement}
The relative-objective configuration improves calibration and rotation,
but worsens depth, camera-center error and raw surfaces. Its lower
validation selection score does not imply lower test error across all
outputs. The composite score encodes a trade-off, and validation covers
only one sequence. Calibration-gradient isolation is a plausible way
to reduce compensation between tasks, but the joint comparison cannot
identify its causal effect. The same applies to relative depth losses,
initial-pose supervision and the refinement penalty.

Refinement raises test ATE in all three CO3D configurations and on TUM.
This establishes that the trained residual update is not uniformly
beneficial under the chosen metric; it does not establish that removing
the module and retraining would yield the same initial-pose scores.
On TUM, its small validation gain and test regression further caution
against selecting a module using the reported test result.
An isolated, predeclared refinement comparison is needed to determine
whether the issue is the update parameterization, supervision or
sequence generalization.

\subsection{Interpreting Auxiliary Objectives}
Observed-view feature distillation is related to predictive representation
learning, but the present task provides target images among the inputs
and therefore does not test withheld-view forecasting.
Correct-versus-shuffled cosine and within-window retrieval
assess target specificity: unstandardized embeddings
can have high cosine with several nearby views. Standardization changes
the feature space, so absolute cosine across configurations is not a
common quality scale. The selected model's near-chance retrieval and
the mixed supplementary ablation results leave a reconstruction benefit
unresolved.

Confidence is similarly target-dependent. An exponential depth-error
score and probability of a relative-depth event should not be compared
numerically as if they represented the same quantity. The latter model
is strongly overconfident on its stated event, and increasing its
filtering threshold does not improve the measured surface scores.
Even a well-calibrated
depth-success probability would not by itself guarantee accurate camera
poses, pointmap residuals or world-coordinate surfaces.

\subsection{Experimental Scope and Limitations}
The primary experiments use one training seed, three CO3D test sequences
from two categories and one indoor test sequence. Windows overlap, and
window-level variation is not uncertainty over independently trained
models or independent scenes. Test results informed successive CO3D
configuration choices; the comparison is adaptive rather than a final
untouched evaluation. We therefore report descriptive paired trends
without treating windows as independent replicates for significance.

CO3D has estimated annotations and dataset-specific scale. TUM is metric,
but the fixed registered-calibration labels limit the intrinsics task.
The CO3D and TUM models are trained separately; the indoor experiment does not
measure zero-shot transfer from objects to rooms. Supplementary component
suites differ in architecture, support, depth representation, view count
or pose initialization and cannot substitute for primary-model ablations.
In particular, random-window validation with noisy pose initialization
addresses a different problem from sequence-held-out RGB-only inference.

Surface recall is measured against observed reference-depth support,
not an independently scanned complete scene. Per-window reconstruction
is not sequence mapping, and observed-view rendering is not novel-view
synthesis. There are no matched primary comparisons with alternative
encoders or state-of-the-art reconstruction systems, no repeated-seed
estimate and no isolated primary-model distillation ablation.
These limits prevent claims of superior reconstruction, a general
pretraining advantage or a causal gain from each component.

\section{Conclusion}
Our architecture adapts dense V-JEPA features to supervised RGB-only
multi-view geometry through explicit cameras, scene aggregation,
epipolar transport and residual pointmaps. Across object-centric and
indoor experiments, camera estimation, depth and aligned shape exhibit
measurable but competing trends. The spatial-camera configuration has
lower motion error than global pooling; the relative-objective
configuration has lower calibration and aligned-shape errors but higher
raw-surface error; and the indoor model has lower camera error than
identity but higher depth error than constant depth. These configuration
comparisons include simultaneous changes in supervision and selection.
The component studies and auxiliary measurements reinforce
that latent agreement or one improved geometric metric is insufficient
to establish consistent reconstruction. The resulting evidence supports
further controlled comparisons of representations and objectives,
with fixed support, explicit alignment and simple geometric baselines.

%% file: appendices/architecture_implementation.tex
\section{Architecture and Training Details}
\label{app:architecture}

The shared encoder is a distilled V-JEPA 2.1 ViT-B/16 initialized from
384-pixel pretraining. At evaluation resolution 512, each independently
encoded image yields 1024 spatial tokens of dimension 768. The local
RGB pyramid has channels 32, 64 and 128. The dense decoder projects
features to 256 channels, combines interpolated edge features at
successive scales and predicts depth, confidence, three normal
coordinates and three point-residual coordinates. The support head is
separate; the descriptor head outputs unit-normalized 64-dimensional
features on the low-resolution grid.

Scene aggregation uses four learned tokens, two transformer layers,
eight attention heads, feed-forward width 1536 and dropout 0.05.
Camera conditioning contains nine rotation entries, three translations
and four normalized intrinsics. The camera refinement head sees the
aggregated feature maps and compares each pooled view with the reference.
The rigid transform obtained from its bounded residual twist
right-multiplies the initial pose. The initial
camera parameterization has no dataset-dependent hard motion bounds.

The spatial camera head projects features to 128 channels on a
$16\times16$ grid. Soft correspondence uses cosine logits at temperature
0.1. The content transformer has two layers and four heads. Its inputs
include the reference feature, expected target feature, their difference,
source and expected target coordinates, cycle coordinates and certainty,
defined as one minus normalized matching entropy.
A second MLP consumes flattened displacement, cycle displacement and
certainty, retaining their spatial arrangement. Calibration is inferred
from pooled reference/view content; the relative-objective configuration
adds a dedicated calibration transformer.

Epipolar transport uses 24 log-spaced depths with endpoints derived
from training data and temperature 0.15.
Candidate points behind the source camera or outside the image do not
contribute. The transported feature and weighted mean candidate depth
are combined with the target feature through convolutional layers.
View interaction is explicit in these modules rather than supplied by
temporal encoding of the independent images.

All primary models train for a 5/10/5-epoch schedule. The first stage
freezes the encoder and disables predicted-camera epipolar transport;
the middle stage adapts the final block and normalization; the final
stage freezes the encoder and increases alignment weights. Evaluation
always enables the configured transport module. The learning rates are
fixed for the respective parameter groups within this schedule.
Automatic mixed precision and activation checkpointing reduce training
memory. Evaluation uses 32-bit floating-point arithmetic. An effective
batch size of four uses two GPUs, one sample per GPU and two accumulation
steps. Each configuration is
trained once with seed 42; no multi-seed uncertainty is available.

The latent predictor has hidden width 512 and a 20-dimensional camera
condition. Teacher targets use an independent frozen copy of the initial
encoder, not an evolving teacher optimized on the reconstruction loss.
Only the relative-objective configuration standardizes targets, using
statistics from 1499 unique training frames and a coordinate standard
deviation floor of 0.005; its 190 validation frames
provide targets without fitting the statistics. The indoor model uses
1915 training and 115 validation target frames without standardization.
Teacher targets are precomputed for reconstruction training and
validation only. Test images are excluded from reconstruction adaptation
and from the fitting of feature-standardization statistics.

%% file: appendices/objective_details.tex
\section{Objective Definitions and Weights}
\label{app:objectives}

Table~\ref{tab:loss-weights} lists the shared objective coefficients
and the changes in the relative-objective configuration.

\subsection{Shared Objectives}
Let $\mathcal V$ denote valid reference pixels, $\ell_{\rm H}$ the
Smooth L1 loss, and
$\rho(x)=\sqrt{x^2+10^{-6}}-10^{-3}$. Local depth uses
$\mathbb E_{\mathcal V}\ell_{\rm H}(\hat d-d)$ and
$\mathbb E_{\mathcal V}e^2-0.85(\mathbb E_{\mathcal V}e)^2$, with
$e=\log\hat d-\log d$. Normal loss averages $1-\hat n^\top n$ where
a complete local depth neighborhood is valid. Depth smoothness weights
finite differences by RGB edges. Intrinsics supervision uses the four
parameters $(f_x/W,f_y/H,c_x/W,c_y/H)$.

Camera translation uses Smooth L1 and rotation uses the geodesic angle
on $SO(3)$. A pose-graph objective supervises pairwise relative transforms
against their reference values; baseline loss compares translation
lengths from the reference camera. Pointmap supervision compares all
three coordinates in camera space. Its residual-magnitude regularizer
is distinct from depth supervision. The shape term fits a similarity
transform to sparse valid world-point correspondences and then averages
robust aligned-coordinate residuals. Gradients pass through the aligned
predictions while holding the fit fixed. Confidence cannot remove
points from this shape loss.

Cross-view visibility uses valid reference depth and the tolerance
$|z-d_j|\leq0.02+0.03d_j$, where $z$ is the projected depth and $d_j$
is the reference depth sampled at the projected target location.
The absolute tolerance uses the respective dataset's depth units.
On visible locations, point consistency
compares predicted world points, and an image cycle compares a
reprojected point with its source coordinate. Photometric and depth
reprojection compare sampled target evidence. Coverage penalizes
projections that leave the image; it supplies a loss even when visible
reprojection support becomes small. Descriptor supervision combines
positive matching, a nonmatch margin and a contrastive term. Spatial
camera correspondence supervision uses bidirectional coarse-grid
cross-entropy with label smoothing 0.05.

Rendering constructs at most 4096 Gaussians from the first view at
pixel stride eight and renders a $96\times96$ RGB/depth target in the
second observed camera. Color comes from source RGB; depth, camera and
normal predictions determine geometry; confidence supplies opacity.
The representation is projective and does not use the pointmap
residual. RGB/depth losses use rendered support and valid target depth
where appropriate, and a coverage term penalizes missing support.

\begin{table*}[t]
\centering\small
\caption{Primary objective coefficients. ``Global/spatial'' denotes the
shared base weighting; the right column specifies the joint
relative-objective configuration. Dashes denote absent terms.}
\label{tab:loss-weights}
\begin{tabular}{lrr}
\toprule
Term & Global/spatial & Relative objectives\\
\midrule
Absolute depth / relative depth / robust log depth & $1.00/---/---$ & $0.25/1.00/0.50$\\
Log-depth residual loss with mean coefficient 0.85 & 0.10 & 0.10\\
Confidence / normals / smoothness & $0.07/0.06/0.04$ & $0.07/0.06/0.04$\\
Intrinsics & 0.035 & 0.50\\
Translation / rotation & $0.18/0.06$ & $0.30/0.06$\\
Initial pose / refinement degradation & $---/---$ & $0.10/0.10$\\
Pose graph / baseline length & $0.10/0.08$ & $0.10/0.08$\\
Camera-frame pointmap / residual magnitude & $0.48/0.025$ & $0.48/0.025$\\
Similarity shape / point consistency / image cycle & $0.48/0.32/0.16$ & $0.48/0.32/0.16$\\
RGB reprojection / depth reprojection / projection coverage & $0.035/0.055/0.01$ & $0.035/0.055/0.01$\\
Rendered RGB / depth / coverage & $0.05/0.05/0.01$ & $0.05/0.05/0.01$\\
Descriptor / contrastive descriptor / support & $0.08/0.02/0.06$ & $0.08/0.02/0.06$\\
Camera matching (spatial configurations only) & 0.10 & 0.10\\
Observed-view distillation & 0.025 & 0.025\\
\bottomrule
\end{tabular}
\end{table*}

\subsection{Relative Geometry and Gradient Paths}
The relative-depth term is
$\mathbb E_{\mathcal V}\rho((\hat d-d)/d)$; robust log-depth uses
Smooth L1 on $\log\hat d-\log d$ with transition 0.1.
Translation supervision divides both prediction and reference by the
window's reference RMS distance of nonreference camera centers from the
first camera. The denominator is lower-bounded by
$\max(0.01d_{\rm near},10^{-4})$, where $d_{\rm near}$ is the
training-depth first percentile.
The initial-pose term supervises translation and rotation before
refinement. The degradation term is
$\max(0,L_{\rm refined}-\operatorname{sg}(L_{\rm initial}))$,
where $\operatorname{sg}$ stops gradients. It compares supervised losses,
not the evaluation ATE directly.

Calibration compares log focal lengths and principal points with
Smooth L1 transitions 0.1 and 0.05, respectively. The normalized
calibration output receives this direct supervision. Its value is used
by camera and geometry modules, but those downstream paths do not
backpropagate into the calibration output. Shared encoder features
remain subject to all applicable losses. Inference uses predicted
calibration throughout.

\subsection{Confidence and Stage Multipliers}
Base confidence is trained with binary cross-entropy to an exponential
absolute-depth-error score, holding depth constant when differentiating
the target. The
relative configuration uses a binary 10\%-relative-error event on valid
pixels and an additional invalid-pixel penalty of weight 0.1. This
changes the target event as well as the loss reduction.
The relative configuration also prevents rendering gradients through
the confidence used as opacity, whereas the base configurations permit
that gradient. These differences are part of the joint intervention.

In stage three, similarity shape, pose graph, point consistency and
image-cycle weights are multiplied by 1.4. Distillation uses stage
multipliers 1.0, 0.75 and 0.50. The relative configuration adds target
retrieval with coefficient 0.25 inside its distillation term. No
coefficient alone establishes relative gradient influence: terms have
different units, scales, reductions and active support.

%% file: appendices/data_protocol_reproducibility.tex
\section{Data and Sampling Details}
\label{app:data}

Tables~\ref{tab:co3d-coverage} and~\ref{tab:tum-split} give the
sequence-disjoint partitions and window counts for the object-centric
and indoor experiments, respectively.

\begin{table*}[t]
\centering\small
\caption{CO3D sequence split. All primary CO3D configurations use the
same sequences and window identities. Each window stores ten sampled
frames; primary evaluation uses five views.}
\label{tab:co3d-coverage}
\begin{tabularx}{\textwidth}{llXr}
\toprule
Partition & Category & Sequence identifiers & Windows\\
\midrule
Train & ball, bench, bowl, broccoli, cake &
ball: 123\_14363\_28981, 375\_42693\_85518;
bench: 415\_57121\_110109;
bowl: 69\_5465\_12831, 70\_5792\_13401;
broccoli: 372\_41112\_81867, 412\_56288\_108844;
cake: 403\_53094\_103680 & 425\\
Validation & bench & 415\_57112\_110099 & 56\\
Test & apple, cake & apple: 110\_13051\_23361, 189\_20393\_38136;
cake: 374\_42274\_84517 & 168\\
\bottomrule
\end{tabularx}
\end{table*}

CO3D frames are ordered within each sequence after removal of frames
without usable depth. Windows sample every fourth retained frame and
advance by three start positions, so many observations are reused.
Images are resized to $512\times512$, with correspondingly transformed
intrinsics. Extrinsics are converted from PyTorch3D axes to the camera
axes used for depth unprojection. Reference depth remains in its
annotation-adjusted dataset scale.

\begin{table*}[t]
\centering\small
\caption{TUM RGB-D sequence-disjoint split. The model is trained
independently of the CO3D models.}
\label{tab:tum-split}
\begin{tabularx}{\textwidth}{lXr}
\toprule
Partition & Sequences & Windows\\
\midrule
Train & fr1/room; fr2/desk; fr2/xyz; fr3/long\_office\_household & 1843\\
Validation & fr1/desk & 97\\
Test & fr3/structure\_texture\_far & 163\\
\bottomrule
\end{tabularx}
\end{table*}

TUM uses timestamp-associated registered depth and interpolated reference
poses. Duplicate pose timestamps are consolidated before interpolation;
RGB/depth separation is at most 20\,ms. The registered images use the
ROS-default calibration and supplied depth correction
\citep{tumformats}; no additional scale correction is applied. Image
letterboxing introduces 64 invalid rows above and below the resized
image. These padding rows are excluded from valid-depth measurements
and predicted surface sampling. Valid depth supplies the support label,
so TUM support is a different task from CO3D object segmentation.

Training-depth statistics set transport endpoints at the first and
99th percentiles and supply the constant-depth control. CO3D near/far
endpoints are 6.6680 and 17.0000, with median
11.4297 in dataset coordinates. TUM endpoints are 0.6230 and 4.9958\,m,
with median 1.5138\,m. These are training-derived statistics, distinct
from the wider accepted depth bounds. They are not fitted to validation
or test depths.

Training samples evenly distributed subsets of each ten-frame window.
Stages one and three sample 3, 5 or 8 views; stage two samples 3 or 5.
A single first-view-relative model is evaluated on a deterministic
five-view subset. The ten-view TUM analysis changes both cardinality
and alignment and is not a controlled cardinality ablation. No primary
experiment reconstructs a single registered map spanning an entire
sequence.

%% file: appendices/metric_definitions.tex
\section{Metric Definitions}
\label{app:metrics}

\subsection{Depth and Pointmaps}
For valid pixels $\mathcal V$ in a window, reference depth $d_p$ and
prediction $\hat d_p$ give
\begin{align}
 \mathrm{RMSE}_D&=\sqrt{\mathbb E_{\mathcal V}(\hat d_p-d_p)^2},\\
 \mathrm{AbsRel}&=\mathbb E_{\mathcal V}\frac{|\hat d_p-d_p|}{d_p}.
\end{align}
Threshold accuracy $\delta_n$ is the fraction with
$\max(\hat d_p/d_p,d_p/\hat d_p)<1.25^n$.
Reference validity uses dataset bounds; the decoder bounds predicted
depth to the same accepted interval. Pointmap RMSE averages squared
coordinate residuals across valid pixels and all three coordinates
before taking a square root, rather than averaging Euclidean norms.
All primary dense metrics are computed within each window and then
averaged over windows.

\subsection{Cameras and Local Trajectories}
Reference poses are expressed in the first-view coordinate frame.
For camera centers $t_k,\hat t_k$, primary ATE is
\begin{equation}
 \mathrm{ATE}=\sqrt{\frac1K\sum_{k=1}^{K}\|\hat t_k-t_k\|_2^2}.
\end{equation}
There is no fitted alignment. Let $\Delta T_k=T_k^{-1}T_{k+1}$,
$\Delta\hat T_k=\hat T_k^{-1}\hat T_{k+1}$ and
$E_k=(\Delta T_k)^{-1}\Delta\hat T_k$.
Translation RPE is the RMS norm of the translation of $E_k$;
rotation RPE is the RMS geodesic angle between the adjacent relative
rotations, expressed in degrees. Intrinsics MAE averages the four
normalized focal-length/principal-point residuals.

The ten-view TUM analysis first fits a rigid rotation and translation
between camera centers, holding scale fixed, and then measures ATE.
Its RPE is computed on adjacent relative poses. Alignment removes the
mean offset of constant identity positions, making the aligned identity
control different from primary raw identity ATE. We report comparisons
only within the same alignment convention.

\subsection{Fixed-Reference Surface Metrics}
For predicted set $P$ and fixed reference set $G$, define
$d(p,G)=\min_{g\in G}\|p-g\|_2$. Accuracy, completeness and Chamfer are
\begin{align}
 a&=|P|^{-1}\sum_{p\in P}d(p,G),\\
 c&=|G|^{-1}\sum_{g\in G}d(g,P),\\
 \mathrm{CD}&=a+c.
\end{align}
Distances are unsquared Euclidean distances. Precision at distance
$\tau$ is the fraction of predicted points within $\tau$ of $G$;
recall reverses the roles. F-score is their harmonic mean. It is
computed per window before averaging and cannot be recovered from
the harmonic mean of aggregate precision and recall.

Reference points are sampled every fourth pixel from valid depth, with
CO3D foreground validity and TUM padding exclusion. The reference set
remains fixed across confidence and support thresholds. Predicted
points can include false positives outside reference validity; these
remain in the precision denominator. Confidence-only conditions threshold
the predicted score; mask conditions also require predicted support
above 0.5. Each set is capped at 50,000 sampled points per window.
Similarity fitting uses at most 20,000 corresponding valid points
before prediction filtering, following least-squares similarity
alignment \citep{umeyama1991least}. This fitted transform is then applied
to predicted points. It is not a transform available at inference.

An empty or nonfinite predicted set is assigned F-score zero and a
Chamfer penalty of $4d_{\max}$, where $d_{\max}$ is the upper accepted
depth bound. Empty-prediction frequencies are reported alongside
F-score at the evaluated confidence thresholds. All primary
nearest-neighbor searches stay within a temporal window; unregistered
windows are not concatenated into one surface.

\subsection{Reliability, Retrieval and Aggregation}
For the 10\%-relative-depth success event, confidence reliability uses
ten equally spaced bins in $[0,1]$. Expected calibration error (ECE)
is the count-weighted absolute difference between mean confidence
and success rate in each bin \citep{guo2017calibration}; binary Brier
score is the mean squared difference between predicted probability
and the binary event label \citep{brier1950verification}.
Pooled-pixel reliability
weights each valid pixel equally; mean window reliability weights each
window equally. They need not coincide. Neither evaluates uncertainty
in camera placement or final world pointmaps.

Target retrieval chooses the largest cosine among the $K$ teacher
embeddings in the same window for each source--target prediction.
With five candidate views, uniform chance is 20\%. Correct-versus-shuffled
cosine uses a cyclic permutation of target identities. These measurements
are evaluated in their respective target feature spaces, not used as a
shared cross-configuration metric when standardization changes.

Primary model selection follows Equation~\ref{eq:selection} for the
spatial and relative-objective configurations and validation AbsRel
for global camera models. Surface threshold is selected separately on
validation raw Chamfer. Earlier exploratory surface sweeps use their
own test-minimum thresholds and pooled support and are labeled as such
in Appendix~\ref{app:additional}.

%% file: appendices/detailed_results.tex
\section{Additional Primary Measurements}
\label{app:detailed-results}

Table~\ref{tab:additional-dense} gives threshold accuracies, pointmap
error and initial/refined camera measurements for all primary models.
Table~\ref{tab:sequence-results} separates the CO3D sequence means,
showing the heterogeneous depth and camera behavior behind the
aggregate. Tables~\ref{tab:complete-surfaces} and~\ref{tab:observed-rendering}
complete the surface and observed-view rendering comparison.

\begin{table*}[t]\centering\small
\caption{Additional dense measurements on the primary five-view windows. TUM distances are meters; CO3D uses dataset units. Pointmap RMSE is coordinate-wise.}
\label{tab:additional-dense}
\begin{tabular}{llrrrrrr}\toprule
Dataset & Configuration & $\delta_2$ & $\delta_3$ & Pointmap RMSE & RPE trans. & Initial ATE & Refined ATE\\\midrule
CO3D & Global camera & 0.9672 & 0.9998 & 1.6767 & 5.5747 & 9.3941 & 9.4225\\
CO3D & Spatial camera & 0.9012 & 0.9993 & 1.6873 & 3.4894 & 6.6350 & 6.6586\\
CO3D & Relative objectives & 0.9161 & 0.9992 & 1.8551 & 3.4961 & 6.8203 & 6.8492\\
TUM & Global camera & 0.2075 & 0.7734 & 0.6288 & 0.1554 & 0.2682 & 0.3063\\
\bottomrule\end{tabular}\end{table*}

\begin{table*}[t]\centering\small
\caption{Sequence-level CO3D depth and camera means. The same sequences are used for all configurations. Identity ATE is independent of the learned model.}
\label{tab:sequence-results}
\begin{tabular}{llrrrr}\toprule
Test sequence & Configuration & Windows & Depth RMSE & AbsRel & ATE\\\midrule
apple/110\_13051\_23361 & Global camera & 56 & 2.1788 & 0.2611 & 10.0505\\
apple/189\_20393\_38136 & Global camera & 56 & 2.9507 & 0.1936 & 13.1926\\
cake/374\_42274\_84517 & Global camera & 56 & 3.3592 & 0.3691 & 5.0244\\
apple/110\_13051\_23361 & Spatial camera & 56 & 3.5735 & 0.4296 & 6.3767\\
apple/189\_20393\_38136 & Spatial camera & 56 & 2.2494 & 0.1409 & 11.1314\\
cake/374\_42274\_84517 & Spatial camera & 56 & 2.8233 & 0.2994 & 2.4675\\
apple/110\_13051\_23361 & Relative objectives & 56 & 2.8970 & 0.3556 & 6.3536\\
apple/189\_20393\_38136 & Relative objectives & 56 & 3.2356 & 0.2136 & 11.5252\\
cake/374\_42274\_84517 & Relative objectives & 56 & 3.4296 & 0.3821 & 2.6688\\
\bottomrule\end{tabular}\end{table*}

These are descriptive window averages, not confidence intervals over independent scenes or training seeds.
The shared evaluation support and window identities permit paired comparisons, while temporal overlap limits statistical independence.

%% file: appendices/surface_render_tables.tex
\begin{table*}[t]\centering\small
\caption{Complete primary surface measurements at validation-selected confidence 0.005. Distances are CO3D coordinates or TUM meters. F-scores use distances 0.05, 0.10 and 0.20 in those units. Each row averages per-window measurements on fixed reference support.}
\label{tab:complete-surfaces}
\setlength{\tabcolsep}{4pt}
\begin{adjustbox}{max width=\textwidth}\begin{tabular}{lllrrrrrr}\toprule
Dataset & Configuration & Alignment & Accuracy & Completeness & Chamfer & F@0.05 & F@0.10 & F@0.20\\\midrule
CO3D & Global camera & Raw & 2.1656 & 1.6212 & 3.7868 & 0.0057 & 0.0170 & 0.0368\\
CO3D & Global camera & Similarity & 0.2718 & 0.3269 & 0.5987 & 0.0854 & 0.2058 & 0.4047\\
CO3D & Spatial camera & Raw & 1.7621 & 1.7914 & 3.5535 & 0.0053 & 0.0157 & 0.0333\\
CO3D & Spatial camera & Similarity & 0.2813 & 0.3635 & 0.6447 & 0.0852 & 0.2045 & 0.3943\\
CO3D & Relative objectives & Raw & 2.0829 & 2.3545 & 4.4374 & 0.0045 & 0.0132 & 0.0265\\
CO3D & Relative objectives & Similarity & 0.2562 & 0.3012 & 0.5573 & 0.0967 & 0.2338 & 0.4476\\
TUM & Indoor model & Raw & 0.5393 & 0.2123 & 0.7515 & 0.0302 & 0.0600 & 0.1191\\
TUM & Indoor model & Similarity & 0.2258 & 0.2934 & 0.5192 & 0.1201 & 0.2267 & 0.4075\\
\bottomrule\end{tabular}\end{adjustbox}\end{table*}

\begin{table*}[t]\centering\small
\caption{Rendering into another observed view. RGB MAE uses intensities in $[0,1]$; depth MAE uses CO3D coordinates or TUM meters. Coverage is image coverage, not reconstructed-scene completeness. Timing includes data loading and geometric measurements in the evaluation loop.}
\label{tab:observed-rendering}
\begin{tabular}{llrrrrr}\toprule
Dataset & Configuration & Windows & RGB MAE & Depth MAE & Coverage & Seconds/window\\\midrule
CO3D & Global camera & 168 & 0.2609 & 2.8848 & 0.9831 & 0.9393\\
CO3D & Spatial camera & 168 & 0.2062 & 2.7765 & 0.9999 & 0.9407\\
CO3D & Relative objectives & 168 & 0.1686 & 3.1039 & 1.0000 & 0.9809\\
TUM & Indoor model & 163 & 0.1627 & 0.9743 & 0.9850 & 0.9440\\
\bottomrule\end{tabular}\end{table*}

%% file: appendices/visual_results.tex
\section{Training Behavior and Multi-View Predictions}
\label{app:visual-results}

\subsection{Validation over the Training Schedule}
\begin{figure*}[t]
\centering
\includegraphics[width=0.96\textwidth]{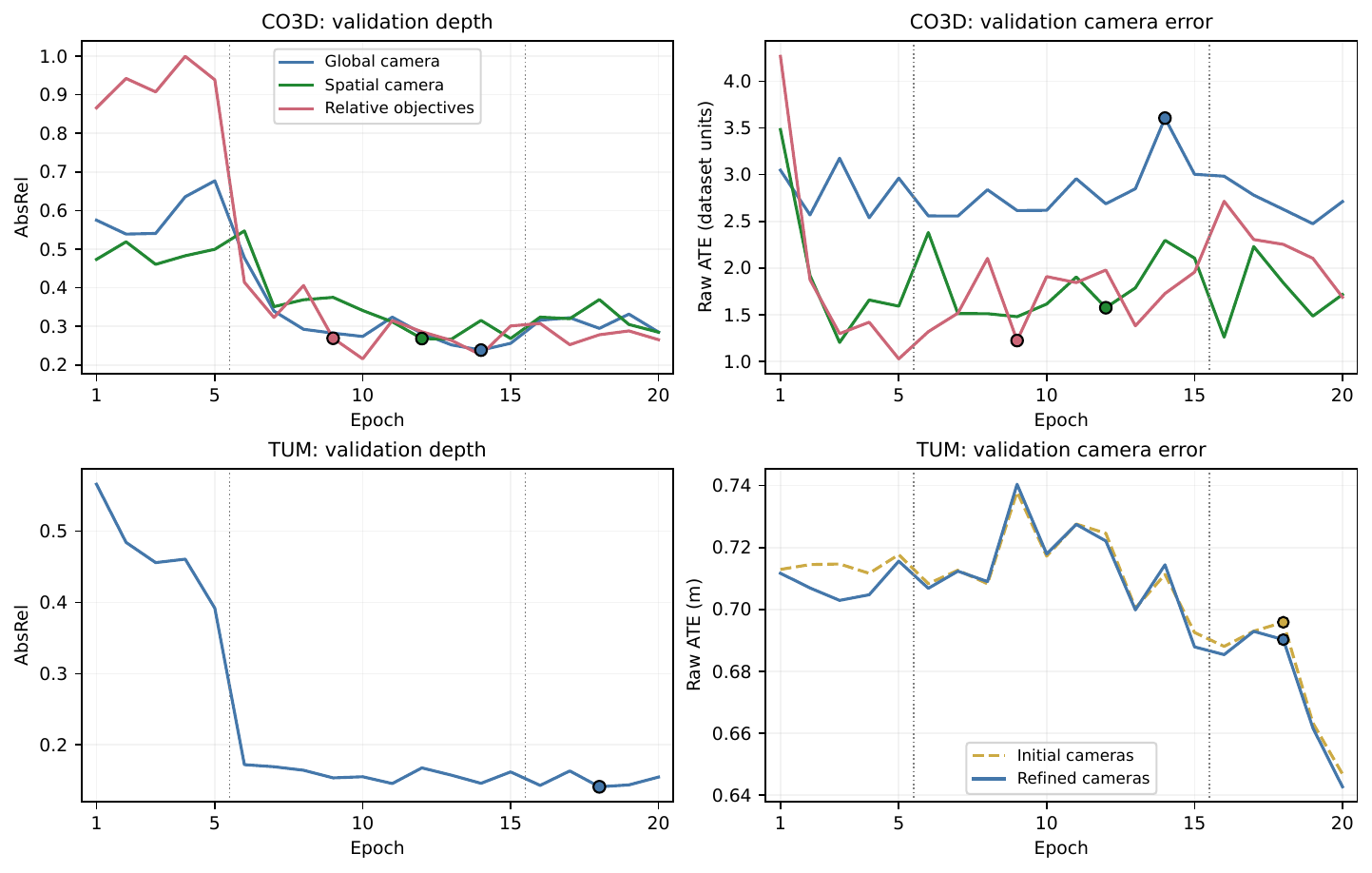}
\caption{Depth and camera validation measurements over 20 epochs.
Markers with black outlines identify the selected epochs: 14, 12 and 9
for global camera, spatial camera and relative objectives on CO3D,
and 18 for the independently trained TUM model. Vertical dotted lines
separate the frozen, partially adapted and alignment-focused stages.
The two TUM camera curves use identical observations and alignment.}
\label{fig:training-generalization}
\end{figure*}

Figure~\ref{fig:training-generalization} shows depth and camera
validation errors across the complete training schedule for both
datasets. All selected epochs precede the end of training. The CO3D
models use their stated, differing selection rules; the curves should
not be interpreted as selection under a common depth-only objective.
On TUM, improving validation depth does not resolve the held-out-sequence
depth gap reported in Section~\ref{sec:tum-results}. The initial/refined
validation camera curves also contrast with refinement's test-time
regression. The validation sequence and objective determine selection;
the test results remain measurements of the selected models.

In the indoor experiment, the weighted projection-coverage penalty
contributes approximately 2.23 to a mean training loss of 2.48 at
epoch 12. Training loss subsequently declines before model selection
at epoch 18. This observation concerns optimization behavior; its
relationship to held-out-sequence depth bias is not established by
an isolated comparison.

\subsection{Depth Predictions across Views}
\begin{figure*}[p]
\centering
\includegraphics[width=0.98\textwidth,height=7.8in,keepaspectratio]{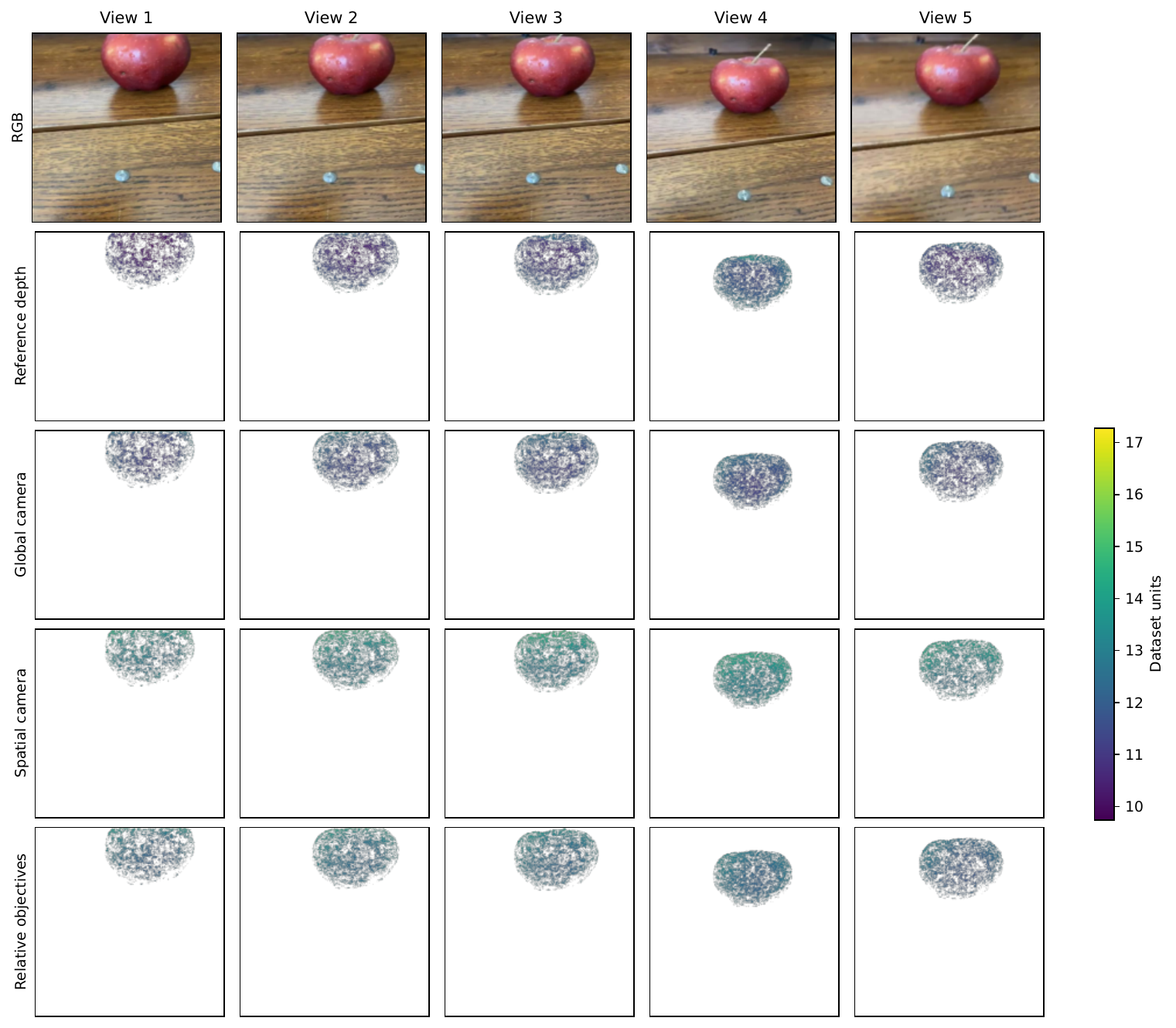}
\caption{CO3D multi-view depth predictions for the first test window.
Rows show RGB, reference depth, global camera, spatial camera and
relative-objective predictions. The RGB observations, reference and
validity mask are identical across configurations. All depth panels use
one scale in CO3D coordinates. White pixels lack valid object-depth
support. The figure illustrates one example, not a ranking over the
entire test set.}
\label{fig:co3d-depth-predictions}
\end{figure*}

\begin{figure*}[p]
\centering
\includegraphics[width=0.98\textwidth,height=7.8in,keepaspectratio]{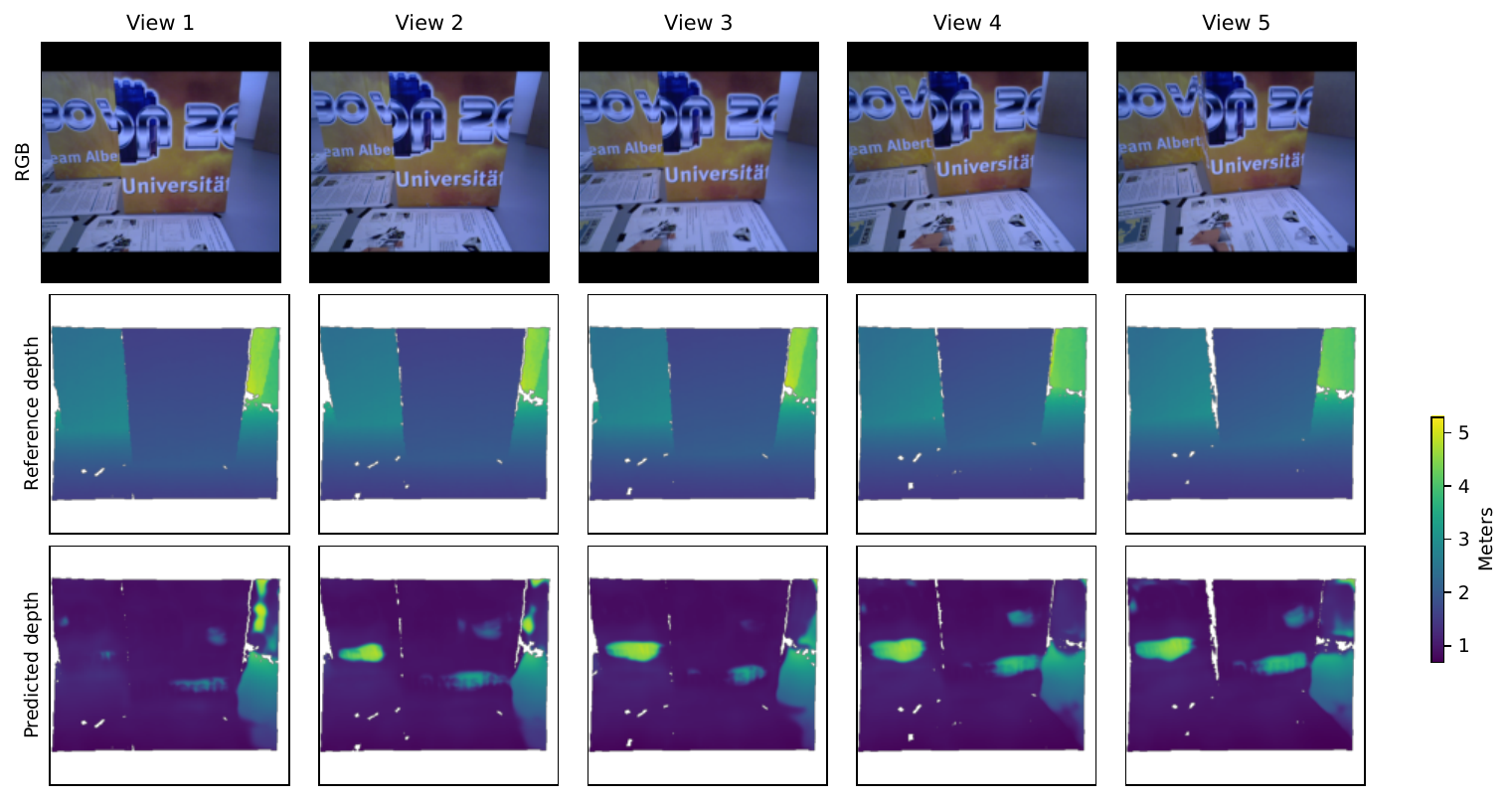}
\caption{TUM multi-view depth predictions for the first test window.
Rows show RGB, reference depth and the independently trained indoor
model's prediction. Depth uses one shared scale in meters across all
views and both depth rows. Invalid reference depth and letterbox
padding are excluded from the depth display.}
\label{fig:tum-depth-predictions}
\end{figure*}

Figures~\ref{fig:co3d-depth-predictions} and~\ref{fig:tum-depth-predictions}
show RGB, reference depth and predicted depth for the same five-view
windows used in the camera and surface illustrations. Within each
figure, one color scale is shared across all depth rows and views.
The CO3D example permits a spatial comparison of all three configurations
on identical observations. On TUM, predicted depth is generally below
the reference through much of the visible scene, consistent with the
signed errors in Figure~\ref{fig:qualitative-depth}. Masking display to
valid reference support makes the depth comparison well-defined; it
does not remove false positives from the independently evaluated
predicted surface. These figures show particular windows, while the
tables summarize all test windows.

\subsection{Confidence Filtering and Empty Predictions}
\begin{figure*}[t]
\centering
\includegraphics[width=0.96\textwidth]{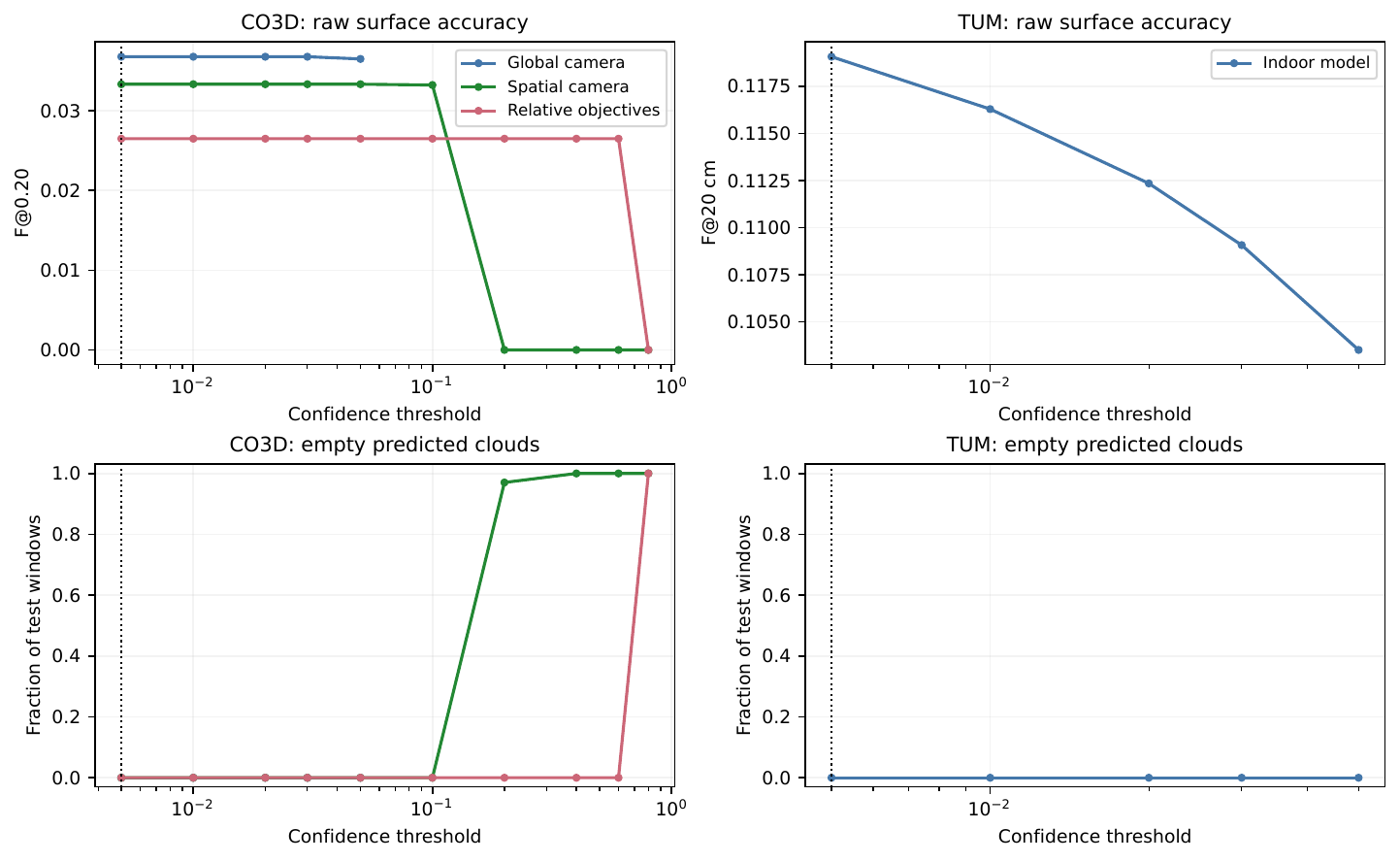}
\caption{Effect of confidence threshold on raw predicted surfaces with
support filtering. Top: mean per-window F-score, using 0.20 CO3D units
or 20\,cm on TUM. Bottom: fraction of windows with empty predicted
clouds. Reference support remains fixed, and empty predictions receive
F-score zero. Vertical dotted lines mark the validation-selected
threshold 0.005; each curve includes only its evaluated thresholds.}
\label{fig:confidence-filtering}
\end{figure*}

Figure~\ref{fig:confidence-filtering} shows surface F-score together
with the fraction of empty clouds. Increasing a threshold can select
too few points without improving surface accuracy. On CO3D, the spatial
camera configuration has 163 empty predictions out of 168 at threshold
0.2; the relative-objective configuration has all predictions empty at
0.8. These failures remain in the averages as zero F-score. On TUM,
the smaller evaluated threshold range already worsens F@20\,cm from
0.1191 to 0.1035. Because the indoor and relative-objective confidence
targets differ, equal numerical thresholds do not define equal
probabilities across models.

%% file: chapters/additional_experiments.tex
\section{Supplementary Component Experiments}
\label{app:additional}

\subsection{Three-View Latent-View Study}
A five-configuration CO3D experiment varies the presence, conditioning
and weight of observed-view distillation. All configurations within
this suite share a 3/6/3-epoch schedule and 195 held-out three-view
windows. Table~\ref{tab:glvp-full} reports their dense and camera
measurements, together with similarity-aligned surfaces at each
configuration's minimum-Chamfer test threshold.

\begin{figure*}[t]
\centering
\includegraphics[width=0.98\textwidth]{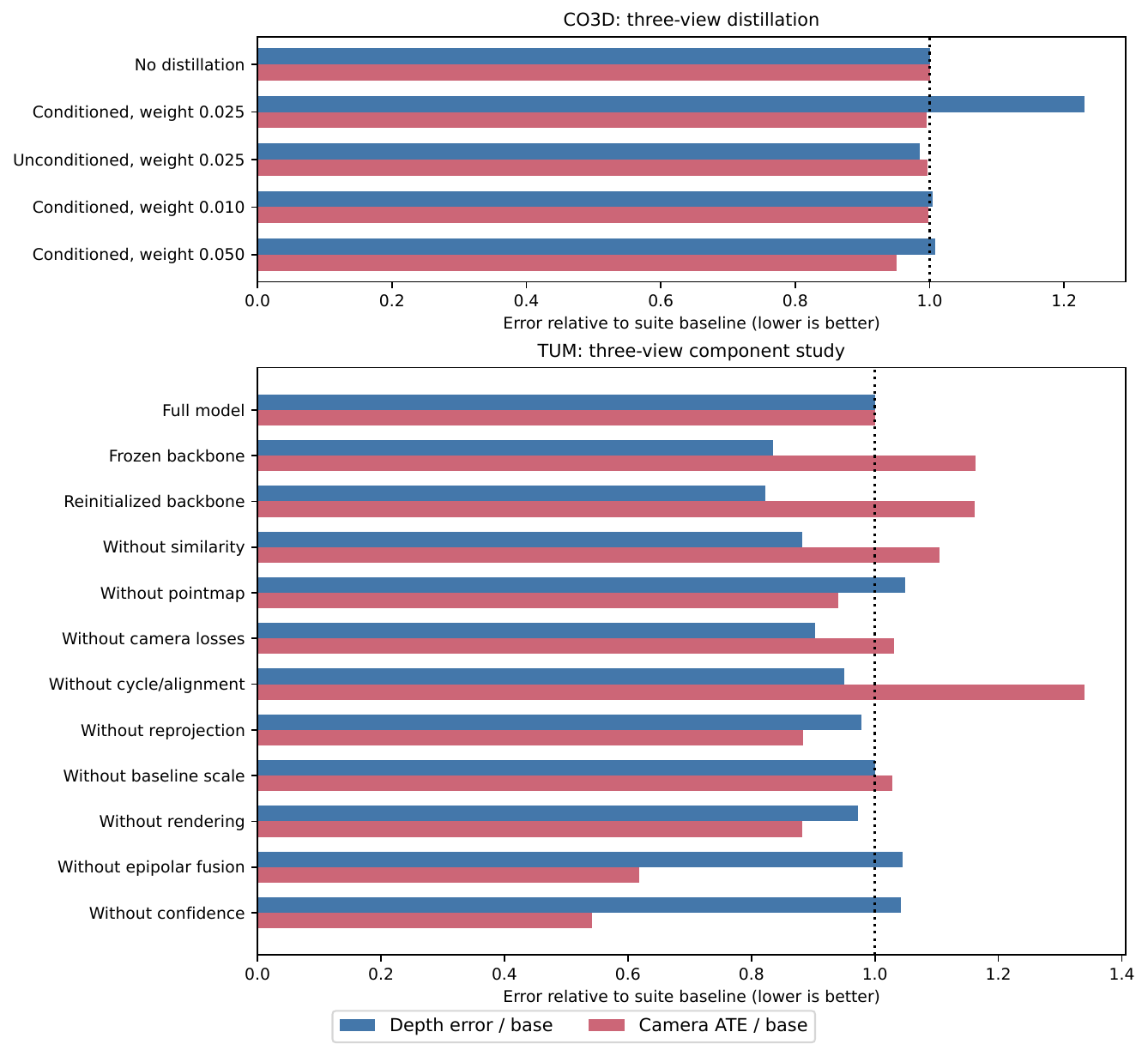}
\caption{Depth and camera trade-offs in the two three-view
component suites. Each metric is divided by its own suite's baseline
error, with lower values better. CO3D measures error against heuristic
numerical depth targets; TUM uses metric depth. These normalized values compare
configurations within a suite, not accuracy across datasets or against
the primary five-view models. The complete unnormalized measurements
are in Tables~\ref{tab:glvp-full} and~\ref{tab:loo-full}.}
\label{fig:component-tradeoffs}
\end{figure*}

The surface protocol differs from the primary experiments in two
material ways. It filters reference support with the prediction and
pools windows expressed in independent local coordinate frames before
nearest-neighbor search. A prediction can therefore affect both the
reference subset and which window supplies a nearest neighbor. The
reported threshold is also selected after examining test surfaces.
Depth targets are formed by selecting, for each image, among arithmetic
rescalings of the stored depth-image values according to the number of
pixels within the accepted depth interval. This differs from the
annotation-based half-precision decoding in the primary experiments.
The depth errors therefore measure agreement with heuristic numerical
targets, not validated depth accuracy under the primary annotation
convention. These conditions prevent comparison of absolute depth or
surface scores with the primary tables. The experiments characterize
within-protocol sensitivity rather than a benchmark ranking or an
ablation of the primary five-view architecture.
Figure~\ref{fig:component-tradeoffs} compares depth and camera changes
relative to the base configuration within each component suite.

\begin{table*}[t]
\centering\small
\caption{Three-view CO3D distillation study. The surface block uses
pooled, prediction-dependent reference support and a test-selected
threshold. Depth-target RMSE measures error against heuristically
rescaled image values. These exploratory values are not interchangeable with the
primary depth and fixed-reference surface metrics.}
\label{tab:glvp-full}
\setlength{\tabcolsep}{4pt}
\begin{adjustbox}{max width=\textwidth}
\begin{tabular}{llrrrrrrr}
\toprule
Distillation & Camera condition & Depth-target RMSE & ATE & RPE rot. & Threshold & Aligned Chamfer & F@0.10 & F@0.20\\
\midrule
None & --- & 0.5153 & 1.7438 & 8.5335 & 0.020 & 0.3171 & 0.6569 & 0.7590\\
Weight 0.025 & Yes & 0.6339 & 1.7359 & 8.3730 & 0.005 & 0.1974 & 0.8604 & 0.9102\\
Weight 0.025 & No & 0.5080 & 1.7388 & 7.5388 & 0.030 & 0.3097 & 0.6023 & 0.7238\\
Weight 0.010 & Yes & 0.5179 & 1.7414 & 16.8072 & 0.030 & 0.7736 & 0.2732 & 0.4040\\
Weight 0.050 & Yes & 0.5196 & 1.6582 & 7.7227 & 0.005 & 0.4009 & 0.6025 & 0.6931\\
\bottomrule
\end{tabular}
\end{adjustbox}
\end{table*}

Adding camera-conditioned distillation at weight 0.025 improves the
selected aligned-surface score while increasing depth-target error.
Removing conditioning gives lower depth-target RMSE but weaker aligned
F-score. Neither smaller nor larger weight yields monotonic gains.
These observations motivate target-discrimination measurements, but
cannot explain the primary model's behavior causally. In particular,
post-hoc threshold selection and altered reference support can amplify
surface differences independently of camera accuracy.

Two further three-view CO3D experiments have different test coverage.
A geometric model evaluated on 260 windows from four sequences has
ATE 1.6967 and rotation RPE $7.8045^\circ$; a model with cached
view targets evaluated on 130 windows from two apple sequences has
ATE 1.4885 and rotation RPE $6.9936^\circ$. Because model and sequence
coverage change together, these rows are not a matched distillation
comparison. Their apparent camera improvement cannot be attributed to
the added objective alone.

An additional five-view object configuration evaluated on 168 windows
has ATE 5.8474 and rotation RPE $17.0449^\circ$. It uses bounded initial
camera motion, heuristically rescaled depth targets and
prediction-selected pooled surfaces. These changes prevent a direct
ranking against the primary configurations, whose initial camera
motion is unbounded and whose surfaces use fixed reference support.

\subsection{Short-Schedule Indoor Component Study}
A separate TUM suite compares the full three-view model with backbone
and loss/module variants using a shared short schedule and held-out
fr3/structure\_texture\_far sequence. Its architecture, view count and
selection objective differ from the primary indoor experiment.
Table~\ref{tab:loo-full} reports all measured configurations. ``Reinitialized''
resets the common learnable backbone layers in that setup; it is not a
primary-model comparison against training an entire alternative encoder
from scratch.

\begin{table*}[t]
\centering\small
\caption{TUM component suite under its shared three-view, short-schedule
protocol. Depth, translation and pointmap errors are in meters; rotation
is in degrees. The aligned pointmap column is a robust coordinate
residual after similarity fitting, not surface Chamfer distance. One
training run is available per configuration.}
\label{tab:loo-full}
\setlength{\tabcolsep}{4pt}
\begin{adjustbox}{max width=\textwidth}
\begin{tabular}{lrrrrrrrr}
\toprule
Configuration & RMSE & AbsRel & $\delta_1$ & ATE & RPE trans. & RPE rot. & Pointmap & Aligned pointmap\\
\midrule
Full model & 0.8570 & 0.3669 & 0.0935 & 0.0811 & 0.0643 & 1.7844 & 0.5344 & 0.2673\\
Frozen backbone & 0.7159 & 0.2882 & 0.2461 & 0.0943 & 0.0730 & 1.8750 & 0.4469 & 0.2224\\
Reinitialized backbone & 0.7051 & 0.2184 & 0.5473 & 0.0942 & 0.0729 & 1.9679 & 0.4470 & 0.1610\\
Without similarity loss & 0.7565 & 0.3030 & 0.2039 & 0.0896 & 0.0697 & 1.9496 & 0.4620 & 0.2018\\
Without pointmap loss & 0.8992 & 0.3756 & 0.0935 & 0.0763 & 0.0616 & 1.7403 & 0.6081 & 0.2948\\
Without camera losses & 0.7734 & 0.3172 & 0.1599 & 0.0836 & 0.0658 & 1.8288 & 0.4787 & 0.2087\\
Without cycle and alignment & 0.8143 & 0.3412 & 0.1249 & 0.1086 & 0.0839 & 2.0585 & 0.5031 & 0.2559\\
Without reprojection & 0.8384 & 0.3614 & 0.1008 & 0.0717 & 0.0590 & 1.7056 & 0.5226 & 0.2328\\
Without baseline-scale loss & 0.8579 & 0.3635 & 0.0793 & 0.0834 & 0.0657 & 1.7825 & 0.5346 & 0.2477\\
Without Gaussian rendering & 0.8335 & 0.3527 & 0.1108 & 0.0715 & 0.0587 & 1.6834 & 0.5179 & 0.1907\\
Without epipolar fusion & 0.8955 & 0.3202 & 0.2595 & 0.0501 & 0.0517 & 1.6774 & 0.5602 & 0.3423\\
Without confidence loss/weighting & 0.8924 & 0.3551 & 0.1432 & 0.0440 & 0.0542 & 1.8011 & 0.5540 & 0.3166\\
\bottomrule
\end{tabular}
\end{adjustbox}
\end{table*}

Frozen and reinitialized backbones lower depth error relative to the
adapted model but raise ATE. Removing epipolar fusion or confidence
supervision lowers ATE while worsening depth and pointmap RMSE. Removing
pointmap supervision gives the largest pointmap error, while removing
cycle and alignment gives the largest camera error. Some removals,
including Gaussian rendering, improve several outputs. Thus the suite
supports task-dependent interactions rather than a consistent benefit
from every loss. Its single-seed outcomes and different model prevent
attributing primary-model gains to those individual modules.

\subsection{Pose-Initialized and Five-View Indoor Settings}
A distinct rendering experiment uses 384-pixel three-view windows,
random-window train/validation separation and noisy pose initialization.
Of 2274 windows, 1932 are used for training and 342 for validation.
The final two encoder blocks are adapted, and the objective includes
RGB/depth Gaussian rendering. Its stage-two validation RMSE is
0.1130\,m, AbsRel 0.0368, $\delta_1$ 0.9891, ATE 0.0980\,m and
rotation RPE $3.8604^\circ$. Temporal overlap and supplied pose
initialization make this a refinement/validation experiment, not a
sequence-held-out RGB-only reconstruction result. Its lower depth
error cannot be taken as evidence that the primary model regressed
under a common protocol.

A separate independently trained five-view indoor model gives RMSE
0.8195\,m, AbsRel 0.3409, ATE 0.1929\,m and pointmap RMSE 0.5102\,m.
Its raw/aligned pooled Chamfer values are 0.1848/0.0890\,m under
prediction-selected reference support. Unlike the primary indoor model,
this model adapts the final two encoder blocks and bounds each initial
translational and rotational twist component to 1.75\,m and 1.05 radians,
respectively. These model and surface-evaluation differences prevent a
controlled ranking against the primary fixed-reference per-window
experiment. The measurements nevertheless illustrate why a large gain
from reference fitting does not establish accurate raw camera placement.

%% file: references.bib
@inproceedings{assran2023ijepa,
  author = {Assran, Mahmoud and Duval, Quentin and Misra, Ishan and Bojanowski, Piotr and Vincent, Pascal and Rabbat, Michael and LeCun, Yann and Ballas, Nicolas},
  title = {{Self-Supervised Learning from Images with a Joint-Embedding Predictive Architecture}},
  booktitle = {{Proceedings of the IEEE/CVF Conference on Computer Vision and Pattern Recognition}},
  year = {2023},
  pages = {15619--15629},
  eprint = {2301.08243},
  archivePrefix = {arXiv},
  url = {https://openaccess.thecvf.com/content/CVPR2023/html/Assran_Self-Supervised_Learning_From_Images_With_a_Joint-Embedding_Predictive_Architecture_CVPR_2023_paper.html}
}

@article{bardes2024vjepa,
  author = {Bardes, Adrien and Garrido, Quentin and Ponce, Jean and Chen, Xinlei and Rabbat, Michael and LeCun, Yann and Assran, Mido and Ballas, Nicolas},
  title = {{Revisiting Feature Prediction for Learning Visual Representations from Video}},
  journal = {Transactions on Machine Learning Research},
  year = {2024},
  eprint = {2404.08471},
  archivePrefix = {arXiv},
  url = {https://openreview.net/forum?id=QaCCuDfBk2}
}

@misc{assran2025vjepa2,
  author = {Assran, Mido and Bardes, Adrien and Fan, David and Garrido, Quentin and Howes, Russell and Komeili, Mojtaba and Muckley, Matthew and Rizvi, Ammar and Roberts, Claire and Sinha, Koustuv and Zholus, Artem and Arnaud, Sergio and Gejji, Abha and Martin, Ada and Robert Hogan, Francois and Dugas, Daniel and Bojanowski, Piotr and Khalidov, Vasil and Labatut, Patrick and Massa, Francisco and Szafraniec, Marc and Krishnakumar, Kapil and Li, Yong and Ma, Xiaodong and Chandar, Sarath and Meier, Franziska and LeCun, Yann and Rabbat, Michael and Ballas, Nicolas},
  title = {{V-JEPA 2: Self-Supervised Video Models Enable Understanding, Prediction and Planning}},
  year = {2025},
  eprint = {2506.09985},
  archivePrefix = {arXiv},
  note = {arXiv preprint arXiv:2506.09985},
  url = {https://arxiv.org/abs/2506.09985}
}

@inproceedings{murlabadia2026vjepa21,
  author = {Mur-Labadia, Lorenzo and Muckley, Matthew and Bar, Amir and Assran, Mido and Sinha, Koustuv and Rabbat, Michael and LeCun, Yann and Ballas, Nicolas and Bardes, Adrien},
  title = {{V-JEPA 2.1: Unlocking Dense Features in Video Self-Supervised Learning}},
  booktitle = {{European Conference on Computer Vision}},
  year = {2026},
  eprint = {2603.14482},
  archivePrefix = {arXiv},
  url = {https://eccv.ecva.net/virtual/2026/poster/4392}
}

@inproceedings{wang2024dust3r,
  author = {Wang, Shuzhe and Leroy, Vincent and Cabon, Yohann and Chidlovskii, Boris and Revaud, Jerome},
  title = {{DUSt3R: Geometric 3D Vision Made Easy}},
  booktitle = {{Proceedings of the IEEE/CVF Conference on Computer Vision and Pattern Recognition}},
  year = {2024},
  pages = {20697--20709},
  eprint = {2312.14132},
  archivePrefix = {arXiv},
  url = {https://arxiv.org/abs/2312.14132}
}

@inproceedings{leroy2024mast3r,
  author = {Leroy, Vincent and Cabon, Yohann and Revaud, Jerome},
  title = {{Grounding Image Matching in 3D with MASt3R}},
  booktitle = {{Proceedings of the European Conference on Computer Vision}},
  year = {2024},
  pages = {71--91},
  doi = {10.1007/978-3-031-73220-1_5},
  eprint = {2406.09756},
  archivePrefix = {arXiv},
  url = {https://www.ecva.net/papers/eccv_2024/papers_ECCV/html/9080_ECCV_2024_paper.php}
}

@inproceedings{wang2025vggt,
  author = {Wang, Jianyuan and Chen, Minghao and Karaev, Nikita and Vedaldi, Andrea and Rupprecht, Christian and Novotny, David},
  title = {{VGGT: Visual Geometry Grounded Transformer}},
  booktitle = {{Proceedings of the IEEE/CVF Conference on Computer Vision and Pattern Recognition}},
  year = {2025},
  pages = {5294--5306},
  eprint = {2503.11651},
  archivePrefix = {arXiv},
  url = {https://arxiv.org/abs/2503.11651}
}

@inproceedings{reizenstein2021co3d,
  author = {Reizenstein, Jeremy and Shapovalov, Roman and Henzler, Philipp and Sbordone, Luca and Labatut, Patrick and Novotny, David},
  title = {{Common Objects in 3D: Large-Scale Learning and Evaluation of Real-Life 3D Category Reconstruction}},
  booktitle = {{Proceedings of the IEEE/CVF International Conference on Computer Vision}},
  year = {2021},
  pages = {10901--10911},
  eprint = {2109.00512},
  archivePrefix = {arXiv},
  url = {https://arxiv.org/abs/2109.00512}
}

@article{kerbl2023gaussians,
  author = {Kerbl, Bernhard and Kopanas, Georgios and Leimk{\"u}hler, Thomas and Drettakis, George},
  title = {{3D Gaussian Splatting for Real-Time Radiance Field Rendering}},
  journal = {ACM Transactions on Graphics},
  volume = {42},
  number = {4},
  year = {2023},
  url = {https://repo-sam.inria.fr/fungraph/3d-gaussian-splatting/}
}

@inproceedings{mildenhall2020nerf,
  author = {Mildenhall, Ben and Srinivasan, Pratul P. and Tancik, Matthew and Barron, Jonathan T. and Ramamoorthi, Ravi and Ng, Ren},
  title = {{NeRF: Representing Scenes as Neural Radiance Fields for View Synthesis}},
  booktitle = {{Proceedings of the European Conference on Computer Vision}},
  year = {2020},
  eprint = {2003.08934},
  archivePrefix = {arXiv},
  url = {https://arxiv.org/abs/2003.08934}
}

@inproceedings{schoenberger2016sfm,
  author = {Sch{\"o}nberger, Johannes L. and Frahm, Jan-Michael},
  title = {{Structure-from-Motion Revisited}},
  booktitle = {{Proceedings of the IEEE Conference on Computer Vision and Pattern Recognition}},
  year = {2016},
  pages = {4104--4113},
  url = {https://openaccess.thecvf.com/content_cvpr_2016/html/Schonberger_Structure-From-Motion_Revisited_CVPR_2016_paper.html}
}

@article{umeyama1991least,
  author = {Umeyama, Shinji},
  title = {{Least-Squares Estimation of Transformation Parameters Between Two Point Patterns}},
  journal = {IEEE Transactions on Pattern Analysis and Machine Intelligence},
  volume = {13},
  number = {4},
  pages = {376--380},
  year = {1991},
  doi = {10.1109/34.88573},
  url = {https://ieeexplore.ieee.org/document/88573}
}

@inproceedings{xu2025freesplatter,
  author = {Xu, Jiale and Gao, Shenghua and Shan, Ying},
  title = {{FreeSplatter: Pose-Free Gaussian Splatting for Sparse-View 3D Reconstruction}},
  booktitle = {{Proceedings of the IEEE/CVF International Conference on Computer Vision}},
  year = {2025},
  pages = {25442--25452},
  eprint = {2412.09573},
  archivePrefix = {arXiv},
  url = {https://openaccess.thecvf.com/content/ICCV2025/html/Xu_FreeSplatter_Pose-free_Gaussian_Splatting_for_Sparse-view_3D_Reconstruction_ICCV_2025_paper.html}
}

@misc{perera2025crossjepa,
  author = {Perera, Avishka and Hewagamage, Kumal and Nazar, Saeedha and Abeywardana, Kavishka and Gallella, Hasitha and Rodrigo, Ranga and Afham, Mohamed},
  title = {{CrossJEPA: Cross-Modal Joint-Embedding Predictive Architecture for Efficient 3D Representation Learning from 2D Images}},
  year = {2025},
  eprint = {2511.18424},
  archivePrefix = {arXiv},
  note = {arXiv preprint arXiv:2511.18424},
  url = {https://arxiv.org/abs/2511.18424}
}

@inproceedings{yang2025fast3r,
  author = {Yang, Jianing and Sax, Alexander and Liang, Kevin J. and Henaff, Mikael and Tang, Hao and Cao, Ang and Chai, Joyce and Meier, Franziska and Feiszli, Matt},
  title = {{Fast3R: Towards 3D Reconstruction of 1000+ Images in One Forward Pass}},
  booktitle = {{Proceedings of the IEEE/CVF Conference on Computer Vision and Pattern Recognition}},
  year = {2025},
  pages = {21924--21935},
  eprint = {2501.13928},
  archivePrefix = {arXiv},
  url = {https://arxiv.org/abs/2501.13928}
}

@inproceedings{cabon2025must3r,
  author = {Cabon, Yohann and Stoffl, Lucas and Antsfeld, Leonid and Csurka, Gabriela and Chidlovskii, Boris and Revaud, Jerome and Leroy, Vincent},
  title = {{MUSt3R: Multi-View Network for Stereo 3D Reconstruction}},
  booktitle = {{Proceedings of the IEEE/CVF Conference on Computer Vision and Pattern Recognition}},
  year = {2025},
  pages = {1050--1060},
  eprint = {2503.01661},
  archivePrefix = {arXiv},
  url = {https://arxiv.org/abs/2503.01661}
}

@inproceedings{liu2025slam3r,
  author = {Liu, Yuzheng and Dong, Siyan and Wang, Shuzhe and Yin, Yingda and Yang, Yanchao and Fan, Qingnan and Chen, Baoquan},
  title = {{SLAM3R: Real-Time Dense Scene Reconstruction from Monocular RGB Videos}},
  booktitle = {{Proceedings of the IEEE/CVF Conference on Computer Vision and Pattern Recognition}},
  year = {2025},
  pages = {16651--16662},
  eprint = {2412.09401},
  archivePrefix = {arXiv},
  url = {https://arxiv.org/abs/2412.09401}
}

@inproceedings{sturm2012tum,
  author = {Sturm, J{\"u}rgen and Engelhard, Nikolas and Endres, Felix and Burgard, Wolfram and Cremers, Daniel},
  title = {{A Benchmark for the Evaluation of RGB-D SLAM Systems}},
  booktitle = {{Proceedings of the IEEE/RSJ International Conference on Intelligent Robots and Systems}},
  year = {2012},
  url = {https://cvg.cit.tum.de/data/datasets/rgbd-dataset}
}

@inproceedings{wang2026pi3,
  author = {Wang, Yifan and Zhou, Jianjun and Zhu, Haoyi and Chang, Wenzheng and Zhou, Yang and Li, Zizun and Chen, Junyi and Pang, Jiangmiao and Shen, Chunhua and He, Tong},
  title = {{$\pi^3$: Permutation-Equivariant Visual Geometry Learning}},
  booktitle = {{International Conference on Learning Representations}},
  year = {2026},
  eprint = {2507.13347},
  archivePrefix = {arXiv},
  url = {https://proceedings.iclr.cc/paper_files/paper/2026/hash/11a09e0aaa74867c6b0719c639fc09f8-Abstract-Conference.html}
}

@inproceedings{lin2026depthanything3,
  author = {Lin, Haotong and Chen, Sili and Liew, Jun Hao and Chen, Donny Y. and Li, Zhenyu and Zhao, Yang and Peng, Sida and Guo, Hengkai and Zhou, Xiaowei and Shi, Guang and Feng, Jiashi and Kang, Bingyi},
  title = {{Depth Anything 3: Recovering the Visual Space from Any Views}},
  booktitle = {{International Conference on Learning Representations}},
  year = {2026},
  eprint = {2511.10647},
  archivePrefix = {arXiv},
  url = {https://proceedings.iclr.cc/paper_files/paper/2026/hash/e4cd50120b6d7e8daff1749d6bbaa889-Abstract-Conference.html}
}

@misc{tumformats,
  year = {n.d.},
  author = {{Technical University of Munich, Computer Vision Group}},
  title = {{TUM RGB-D Dataset: File Formats and Intrinsic Camera Calibration}},
  howpublished = {Online dataset documentation},
  url = {https://cvg.cit.tum.de/data/datasets/rgbd-dataset/file_formats},
  note = {Accessed November 5, 2026}
}

@article{ye2025gsplat,
  author = {Ye, Vickie and Li, Ruilong and Kerr, Justin and Turkulainen, Matias and Yi, Brent and Pan, Zhuoyang and Seiskari, Otto and Ye, Jianbo and Hu, Jeffrey and Tancik, Matthew and Kanazawa, Angjoo},
  title = {{gsplat: An Open-Source Library for Gaussian Splatting}},
  journal = {Journal of Machine Learning Research},
  volume = {26},
  number = {34},
  pages = {1--17},
  year = {2025},
  url = {https://www.jmlr.org/papers/v26/24-1476.html}
}

@inproceedings{loshchilov2019adamw,
  author = {Loshchilov, Ilya and Hutter, Frank},
  title = {{Decoupled Weight Decay Regularization}},
  booktitle = {{International Conference on Learning Representations}},
  year = {2019},
  url = {https://openreview.net/forum?id=Bkg6RiCqY7}
}

@misc{oord2018cpc,
  author = {van den Oord, Aaron and Li, Yazhe and Vinyals, Oriol},
  title = {{Representation Learning with Contrastive Predictive Coding}},
  year = {2018},
  eprint = {1807.03748},
  archivePrefix = {arXiv},
  note = {arXiv preprint arXiv:1807.03748},
  url = {https://arxiv.org/abs/1807.03748}
}

@inproceedings{eigen2014depth,
  author = {Eigen, David and Puhrsch, Christian and Fergus, Rob},
  title = {{Depth Map Prediction from a Single Image using a Multi-Scale Deep Network}},
  booktitle = {{Advances in Neural Information Processing Systems}},
  volume = {27},
  year = {2014},
  url = {https://proceedings.neurips.cc/paper_files/paper/2014/hash/91c56ce4a249fae5419b90cba831e303-Abstract.html}
}

@inproceedings{guo2017calibration,
  author = {Guo, Chuan and Pleiss, Geoff and Sun, Yu and Weinberger, Kilian Q.},
  title = {{On Calibration of Modern Neural Networks}},
  booktitle = {{Proceedings of the 34th International Conference on Machine Learning}},
  series = {Proceedings of Machine Learning Research},
  volume = {70},
  pages = {1321--1330},
  year = {2017},
  url = {https://proceedings.mlr.press/v70/guo17a.html}
}

@article{brier1950verification,
  author = {Brier, Glenn W.},
  title = {{Verification of Forecasts Expressed in Terms of Probability}},
  journal = {Monthly Weather Review},
  volume = {78},
  number = {1},
  pages = {1--3},
  year = {1950},
  url = {https://doi.org/fp62r6}
}
